%% file: main.tex
\documentclass[journal,final]{IEEEtran}

\input{preamble}
\usepackage{pdfpages}
\usepackage[dvipsnames]{xcolor}
\definecolor{blue}{rgb}{0,0,1.0}
\definecolor{green}{rgb}{0.5,1.0,0.5}
\usepackage{xr-hyper}
\usepackage[pagebackref,breaklinks,colorlinks,citecolor=blue]{hyperref}
\usepackage[normalem]{ulem}
\newcommand{\siamul}[1]{{\color{black}#1}} 
\newcommand{\siamulnew}[1]{{\color{black}#1}} 
\newcommand{\siamulnewnew}[1]{{\color{black}#1}} 
\newcommand{\siamulnewnewadded}[2]{%
  {\color{ForestGreen}\bgroup\markoverwith{\textcolor{#1}{\rule[-0.5ex]{3pt}{1.0pt}}}\ULon{#2}}%
}

\newcommand{\siamulnewnewdeleted}[2]{%
  {\color{red}\bgroup\markoverwith{\textcolor{#1}{\rule[0.5ex]{3pt}{1.5pt}}}\ULon{#2}}%
}

\usepackage{mathtools}
\usepackage{bm}
\usepackage{boldline}
\usepackage{enumitem}

\title{EyePreserve: Identity-Preserving Iris Synthesis}

\author{Siamul~Karim~Khan,~\IEEEmembership{Student Member,~IEEE,}
        Patrick~Tinsley,~\IEEEmembership{Student Member,~IEEE,} \\
        Mahsa~Mitcheff,~\IEEEmembership{Student Member,~IEEE,}
        Patrick~J.~Flynn,~\IEEEmembership{Fellow,~IEEE,} \\
        Kevin~W.~Bowyer,~\IEEEmembership{Fellow,~IEEE,}
        and~Adam~Czajka,~\IEEEmembership{Senior~Member,~IEEE}
\thanks{The authors are with the Department of Computer Science and Engineering, University of Notre Dame, Indiana, USA. Corresponding e-mail: skhan22@nd.edu. 

\copyright~2026 IEEE. Personal use of this material is permitted. Permission from IEEE must be obtained for all other uses, in any current or future media, including reprinting/republishing this material for advertising or promotional purposes, creating new collective works, for resale or redistribution to servers or lists, or reuse of any copyrighted component of this work in other works.}
}

\begin{document}



\maketitle

\ifCLASSOPTIONpeerreview
\begin{center} \bfseries EDICS Category: 3-BBND \end{center}
\fi
%
\IEEEpeerreviewmaketitle

\input{sections/0_abstract}    
\input{sections/1_intro}
\input{sections/2_relatedwork}

\input{sections/3_data}
\input{sections/4_iris_texture_deformation_model}
\input{sections/5_results}

\input{sections/6_limitations}
\input{sections/7_conclusions}


\bibliographystyle{IEEEtran}
\bibliography{ref}

\input{biographies}

\end{document}

%% file: preamble.tex
\usepackage{caption}
\usepackage{verbatim}
\usepackage[dvipsnames]{xcolor}
\usepackage{multirow}
\usepackage{float}
\usepackage{adjustbox}
\usepackage{cuted}
\usepackage{stfloats}
\usepackage{siunitx,etoolbox}
\usepackage{subcaption}
\usepackage{times}    
\usepackage{cite}     
\usepackage{xspace}
\usepackage{graphicx}
\usepackage{amsmath}
\usepackage{amssymb}
\usepackage{booktabs}
\usepackage[numbers,sort&compress]{natbib}
\usepackage{silence}
\usepackage{etoolbox}
\usepackage{keyval}
\usepackage{color}
\usepackage{float}
\usepackage[compact]{titlesec}
\titlespacing*{\section}{0pt}{12pt plus 2pt minus 2pt}{4pt plus 1pt minus 1pt}
\titlespacing*{\subsection}{0pt}{10pt plus 2pt minus 2pt}{3pt plus 1pt minus 1pt}
\usepackage{needspace}
\makeatletter
\DeclareRobustCommand\onedot{\futurelet\@let@token\@onedot}
\def\@onedot{\ifx\@let@token.\else.\null\fi\xspace}

\def\eg{\emph{e.g}\onedot} 
\def\ie{\emph{i.e}\onedot}

\def\etal{\emph{et al}\onedot}

\makeatother

%% file: sections/0_abstract.tex
\begin{abstract}

Synthesis of same-identity biometric iris images, both for existing and non-existing identities, while preserving the identity across a wide range of pupil sizes, is complex due to the intricate iris muscle constriction mechanism, requiring a precise model of iris non-linear texture deformations to be embedded into the synthesis pipeline. \siamulnew{This paper presents {\it EyePreserve}, a novel, fully data-driven non-linear texture deformation model, embedded within an identity-preserving, pupil size-varying synthesis framework.} This approach is capable of synthesizing images of irises with different pupil sizes representing non-existing identities, as well as non-linearly deforming the texture of iris images of existing subjects, given the segmentation mask of the target iris image. Iris recognition experiments suggest that the proposed deformation model both preserves the identity when changing the pupil size, and offers better similarity between same-identity iris samples with significant differences in pupil size, compared to state-of-the-art linear and non-linear (biomechanical-based) iris deformation models. Two immediate applications of the proposed approach are: (a) synthesis \siamul{of new,} or enhancement of the existing datasets \siamul{of same-identity} iris \siamul{images with varying pupil sizes, with correct modeling of complex iris texture deformations}, and (b) helping forensic human experts examine iris image pairs with significant differences in pupil dilation \siamul{by deforming one or both images to align the pupil size}. Images considered in this work conform to selected ISO/IEC 29794-6 quality metrics to make them applicable in biometric systems. The source codes and model weights are offered with this paper.

\end{abstract}


%% file: sections/1_intro.tex
\section{Introduction}
\label{sec:intro}

\begin{figure*}[!ht]
\centering
\includegraphics[width=0.92\linewidth]{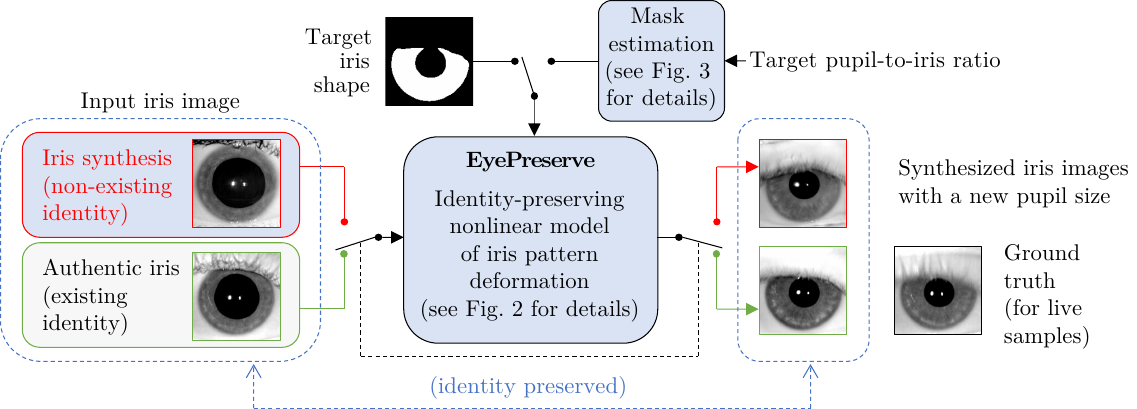}
\setlength{\belowcaptionskip}{-7pt}
\caption{{\it EyePreserve} accepts (a) an iris image (either synthetically generated or of a non-existing identity, or authentic of an existing subject) and (b) a target shape of the iris expressed either as a mask or a new pupil-to-iris ratio. By combining (a) and (b), the {\it EyePreserve} synthesizes a new iris image with the iris texture deformed to match a given new iris shape. The proposed model preserves the identity and correctly models non-linear deformations of the iris muscle.}
\label{fig:overview}
\end{figure*}

Effective same-identity biometric sample generation relies on algorithms that successfully balance visual realism with the preservation of identity-specific characteristics. While older algorithms (non-deep-learning-based)~\cite{Makthal_EUSIPCO_2005} and modern generative models~\cite{Yadav_CVPRW_2019,yadav2023iwarpgan} can synthesize realistic-looking irises, they struggle to maintain biometric identity across different physiological states. Specifically, modeling complex structural changes, such as the non-linear way in which iris textures deform during pupil dilation, remains a critical research gap. Existing solutions often depend on biomechanical assumptions that fail under extreme dilation, or they rely on linear scaling models (\eg, Daugman's ``rubber sheet'' model) that introduce significant geometric noise. The preliminary {\it DeformIrisNet} solution \cite{khan2023deformirisnet} provided a valuable proof-of-concept as the first autoencoder-based approach for iris texture deformation. However, it depended on static loss penalties, which led to overfitting. To overcome these limitations, we fundamentally restructure the optimization landscape and introduce {\it EyePreserve}, an identity-preserving mechanism that synthesizes same-eye biometric iris images with varying pupil sizes for both non-existing identities and existing iris samples. To effectively anchor identity features during texture deformation, our proposed framework introduces a novel loss function that integrates filter-based, triplet margin, and adversarial components. Rather than relying on static physical formulas, our autoencoder-based model takes an iris image and a target shape mask as inputs, and is guided by this specialized loss formulation to learn the biology-driven deformations of iris muscle fibers directly from live iris near-infrared videos. This approach eliminates the need for prior biomechanical assumptions while guaranteeing identity preservation when deforming the iris texture. By operating on top of a fully differentiable pipeline, our adversarial triplet formulation enforces a metric learning space, actively pushing inter-class identities apart while pulling intra-class samples together, while focusing exclusively on the iris texture. This explicit enforcement of biometric boundaries yields a conceptually distinct model that generalizes robustly across diverse datasets. Experiments demonstrate that \siamul{{\it EyePreserve}} offers better compensation for pupil size variations compared to the linear model~\cite{daugman1993high}, non-linear bio-mechanical model \cite{tomeo2015biomechanical}, and {\it DeformIrisNet}. Figure~\ref{fig:overview} provides an overview of how the proposed {\it EyePreserve} method works. 

\siamulnew{
In summary, the {\bf core conceptual and technical contributions} of the {\it EyePreserve} framework are as follows:
\begin{enumerate}[leftmargin=5.2mm]
\item[(a)] A robust, non-linear iris texture deformation model that formulates iris biomechanics as a mask-guided, image-to-image translation problem. 
\item[(b)] A novel identity-preserving loss function that uniquely combines filter-based, triplet margin, and adversarial components to better preserve identity during iris texture deformation. This enables successful generalization across diverse datasets--a key limitation of prior deep learning models--and yields superior recognition performance compared to standard linear deformation models.
\item[(c)] A comprehensive, end-to-end synthesis framework that seamlessly integrates high-fidelity iris generation models with our novel identity-preserving deformation mechanism, offering a complete tool for generating perceptually realistic, pupil-size-varying iris images without compromising biometric utility.
\end{enumerate}
}
The source codes (including the training routines) and all model weights are offered with this paper to facilitate replicability and various applications of this work: \url{https://github.com/CVRL/EyePreserve}.

%% file: sections/2_relatedwork.tex
\section{Related Work}
\label{sec:relatedwork}
\subsection{Iris Texture Deformation}

Methods have been suggested for linear and non-linear iris texture deformation, which align iris texture patterns accounting for elastic deformations of the iris muscles.
In Daugman’s ``rubber sheet'' model~\cite{daugman1993high}, the annular iris region is stretched to a fixed-sized rectangular block to account for pupil size changes. The limitation of this linear model is that significant pupil dilation can result in severe deviations from linear movements of the iris pattern. The Wyatt model~\cite{wyatt2000minimum} is a non-linear model based on Rohen's meshwork~\cite{rohen1951bau}\footnote{The iris collagen structure is organized in a series of parallel fibers (arcs) connecting the pupil margin to the iris root (boundary) in clockwise and counterclockwise directions of 90 degrees.} is designed to minimize iris stretching as the pupil size changes.
This is achieved by determining the optimal properties of the slopes between two arcs (fibers) and allowing points on a fiber to move only in the radial direction. The model introduced by Yuan and Shi~\cite{yuan2005non} builds upon Wyatt's iris fiber structure by combining both linear and non-linear techniques and correcting scaling problems resulting from variations in distance. Reyes \etal~\cite{tomeo2015biomechanical} introduced a biomechanical non-linear normalization where the iris is conceptualized as a thin cylindrical shell of orthotropic material, and the model calculates iris deformation in the radial direction using stress and strain vectors. 

The above models were created to be part of the iris recognition pipeline, rather than to serve as identity-preserving elements of an iris synthesis pipeline. More recently, Khan \etal~\cite{khan2023deformirisnet} proposed an autoencoder-based iris texture deformation model, developed and tested on iris images from a single source. Although this model was the first end-to-end, fully deep learning-based model of iris texture deformation, \siamul{it was developed and tested on subject-disjoint iris images but only from a single source, \ie,} its identity preservation in a cross-dataset setting was not examined. \siamul{Our experiments reveal that DeformIrisNet performs worse when tested on datasets other than Warsaw-Biobase Pupil Dynamics (WBPD)~\cite{Kinnison_ICB_2019}, indicating overfitting to training data characteristics. The proposed {\it EyePreserve} model significantly improves upon DeformIrisNet by incorporating a more advanced triplet-adversarial loss and by demonstrating strong cross-dataset generalization capabilities.}

\subsection{Synthetic Iris Image Generation}

\siamul{Recent work has explored iris synthesis under variable acquisition conditions.} Yadav \etal\cite{Yadav_CVPRW_2019} applied a relativistic standard GAN (RaSGAN) to synthesize iris images used later to train iris presentation attack detection methods. This approach, however, did not offer control over the properties of synthesized irises, including the identity-related features. In ~\cite{yadav2023iwarpgan}, iWarpGAN was introduced to synthesize iris images not seen during training. It utilizes two distinct input images with different identities and styles. The generated image has a unique identity from the first input image and adopts the style of the second input image. Its limitation is that the number of new synthetic identities is the same as the number of real identities in the dataset. \siamul{Boutros \etal~\cite{boutros2020iris} proposed a two-stage D–ID-Net: a domain network (D-Net) that maps semantic segmentation masks to generic eye images, followed by an identity-specific network (ID-Net), trained per identity, that injects identity into the D-Net outputs. Identity preservation is evaluated by comparing verification performance on D-Net vs. ID-Net outputs using the OpenEDS dataset. Kakani \etal~\cite{kakani2024segmentation} present a SegNet–IDNet–IrisGAN pipeline that conditions a GAN generator on segmentation masks and identity features to synthesize higher-fidelity, identity-preserving iris images. Their experiments are also performed on OpenEDS, and the authors note the need for cross-dataset validation. \siamulnew{Li \etal proposed I3FDM~\cite{li2024i3fdm}, a method that uses diffusion models to realistically fill in blocked or masked portions of iris images. By generating smooth, plausible textures to replace these disruptions, their approach prevents structural errors during image processing, which ultimately improves overall iris recognition accuracy.} 

\siamulnew{Expanding beyond tightly cropped iris generation, many of the recent research works focused on synthesizing the entire ocular region to capture broader contextual features and cross-spectral variations. Toma{\v{s}}evi{\'c} \etal~\cite{tomavsevic2022bioculargan} introduced BiOcularGAN, a dual-branch framework capable of synthesizing aligned bimodal ocular images (visible and near-infrared) alongside corresponding semantic segmentation masks extracted directly from the model's latent features. Similarly, to utilize the broader ocular context for specialized security tasks, Yadav and Ross~\cite{yadav2025multi} proposed MID-StyleGAN, a multi-domain translative framework combining diffusion models and GANs to synthesize high-fidelity ocular images capturing both bona fide and presentation attack characteristics.}

In general, current generative models can change iris image styles, but most of them do not focus on preserving identity, and none of them, to our knowledge, have been shown to generalize across datasets with different acquisition conditions. In contrast, in this paper, we investigate identity preservation under varying pupil sizes, across both real and synthetic data sources, and across multiple iris datasets. } 

\subsection{Identity Preservation in \siamul{Other Biometric Modalities}}

\siamulnew{Although our primary focus is on iris images, the challenge of identity-preserving synthesis is deeply rooted in broader facial generation research as well, albeit with face domain-specific complexities. Historically, however, generative models in both domains have prioritized subjective visual realism over strict biometric utility.}

\siamulnew{A significant portion of foundational research in this domain focused on disentangling and navigating the latent spaces of pre-trained GANs to achieve identity-preserving synthesis.} Tzelepis \etal~\cite{tzelepis2021warpedganspace} introduced WarpedGANSpace, a method of face synthesis that aims to modify face styles by exploring radial basis function paths in a GAN's latent space. The absence of matching scores for the different face styles does not allow us to fully assess the fidelity of identity preservation offered by this model. Shoshan \etal~\cite{shoshan2021gan}, Liu \etal~\cite{liu2022controllable}, and Li \etal~\cite{li2022cross} present controllable GAN models that are capable of generating face images of varying styles while preserving identity. Their models were developed in order to augment face recognition datasets to improve recognition accuracy and generalizability. FaceID-GAN~\cite{shen2018faceid} accomplishes the same task by adding a third model (a classifier component) to the GAN training process to encourage same-subject image synthesis.  Other works~\cite{zhu2015high, yin2017towards} do not focus on synthesizing biometric data {\em per se}, but rather on re-styling authentic in-the-wild faces to fully frontal, neutral expression images. IPGAN~\cite{yan2022ipgan} tries to preserve facial identity while creating caricature faces. Later, Li \etal~\cite{li2022cross} use a similar concept to maintain the identity and pose of a face in an input image while translating it to an image showing the face with different appearances and styles. 

\siamulnew{More recently, diffusion models and latent space autoencoders have significantly advanced identity-preserving face generation. Kim \etal introduced DCFace~\cite{kim2023dcface}, a dual-condition model combining subject ID with external styles to improve downstream face recognition accuracy. Boutros \etal proposed IDiff-Face~\cite{boutros2023idiff}, an identity-conditioned latent diffusion model utilizing Contextual Partial Dropout (CPD) to balance identity discrimination with realistic intra-class variation. Toma{\v{s}}evi{\'c} \etal developed ID-Booth~\cite{tomavsevic2025id}, leveraging a novel triplet training loss with a pre-trained face recognition network for privacy-preserving, identity-consistent dataset augmentation. Taking a text-free approach, Papantoniou \etal introduced Arc2Face~\cite{papantoniou2024arc2face}, a foundation model conditioned entirely on ArcFace features to generate diverse, photo-realistic faces. Finally, pushing dataset scale, Wu \etal introduced Vec2Face~\cite{wu2025vec2face}, which perturbs identity vectors through a feature masked autoencoder to synthesize up to 300K distinct identities with natural intra-class variations, rivaling real-world training data.}

While identity preservation has been studied in face generation, iris image generation \siamulnew{presents a different challenge} 
as the identity information is embedded across a range of frequencies compared to the low-frequency facial features.

%% file: sections/3_data.tex
\section{Datasets}
\label{sec:data} 

The existing ISO/IEC 29794-6-compliant iris datasets usually lack samples acquired under intentional ambient light changes to control pupil dilation. Thus, we acquired samples from four distinct sources, as outlined in the following subsections, to enhance the {\it EyePreserve} model's ability to learn non-linear deformations and to provide comprehensive subject-disjoint evaluations of the proposed model. \siamulnew{Our data selection focuses on standard-quality (\ie compliant with ISO/IEC 29794-6) images, but with challenging pupil dilation differences. While there are many unconstrained iris and ocular image datasets~\cite{de2015mobile, tonsen2016labelled}, these images are usually dominated by confounding factors, such as insufficient lighting, reflections, or off-angle occlusions, making it difficult to focus on the intricate changes in iris texture. Omitting these variables ensures our evaluation strictly measures the model's effectiveness on structural iris deformations, rather than its robustness to poor acquisition conditions.}

\subsection{Training Datasets}

\subsubsection{Data Sources}

\textbf{Warsaw BioBase Pupil Dynamics (WBPD)}~\cite{kinnison2019learning} is composed of 159 high-resolution ($768\times576$ px) iris videos from 42 individuals' eyes under varying lighting conditions (117,717 iris images in total). Each subject's eyes were captured at a 25 FPS rate for 30 seconds: the first 15 seconds in darkness, the next 5 seconds in elevated light intensity (pupil constriction period), and the remaining 10 seconds in decreased light intensity (pupil dilation period). \textbf{CSOSIPAD} is a subset of the ``Combined Dataset'' introduced by Boyd \etal in~\cite{BXGRID} that was used in conjunction with the WBPD dataset to inject more diversity into the training set for the models. Whereas WBPD represents 84 different irises (42 subjects $\times$ 2 eyes), CSOSIPAD added 1,627 distinct irises (each iris is considered as a different identity).


\subsubsection{WBPD-specific curation} Iris images in which the iris texture was not visible \siamul{(\eg, due to blinking)} were excluded from training. Furthermore, in order to avoid identity leakage (as described in \eg \cite{tinsley2022haven}) and redundancy of very similar images seen during training, images in seconds 0-14 of each video were also excluded. After these two exclusions, 55,947 images remained.

\subsubsection{CSOSIPAD-specific curation} Samples collected by the LG2200 sensor, which showed interlace artifacts, were removed. The remaining 50,167 CSOSIPAD images were centrally cropped around the iris in the same manner as the WBPD data. 

\subsubsection{Curation common for WBPD and CSOSIPAD} \siamulnew{First, the iris images were tightly cropped around the iris bounding box to a spatial resolution of $256\times256$ pixels. Because this crop is tight ($\frac{16}{14}$ times the iris radius on each side), the resulting iris diameter within these boundaries is approximately 224 pixels. This safely exceeds the 160-pixel minimum diameter required by the ISO/IEC 29794-6 standard for high-quality biometric enrollment samples. Consequently, these parameters define the spatial dimensions and iris diameter of the synthesized outputs.} 

\siamul{Training the {\it EyePreserve}} model also requires accurate iris segmentation masks, which we generated using Vance \etal's models~\cite{vance2022deception}. Finally, to train the deformation model effectively, we needed appropriate input-target pairs featuring the same eye with varying pupil sizes. We first calculated the pupil and iris radii for all images in the WBPD and CSOSIPAD datasets. Following suggestions from~\cite{hollingsworth2009pupil, tomeo2015biomechanical}, we isolated images with a pupil-to-iris ratio between 0.2 and 0.7. Then, we divided these images into five bins of width 0.1 (\eg, the first bin contains images with ratios between 0.2 and 0.3). Finally, to construct our training pairs, images in each bin were paired with images from all other bins, strictly ensuring that every pair corresponds to the same unique identity (matching both the subject ID and the specific eye type).

\subsection{Test Datasets}
\label{sec:testdata}

\siamul{The first test subset is the combination of (a) an eye-disjoint subset of \textbf{WBPD} representing data from three eyes, and (b) a newly-collected dataset from five unique eyes that is offered with this paper. We collectively call this test set WBPD-plus.}

The second test set,
the \textbf{Hollingsworth Split (HWS)} dataset~\cite{hollingsworth2009pupil} contains images of irises with pupil dilation variations for the same identities and was used to demonstrate that pupil dilation can degrade iris biometric performance. This dataset is an eye-disjoint subset of CSOSIPAD. 

The third test set, the \textbf{Quality-Face/Iris Research Ensemble (Q-FIRE)} dataset~\cite{QFIRE}, contains near-infrared (NIR) videos of individuals at different distances and illumination levels. We extract iris images from Q-FIRE videos, and \siamulnew{the source code for this extraction from the Q-FIRE videos is provided with this paper for reproducibility}.

\siamulnew{The fourth test set, the Warsaw-BioBase-SmartPhone-Iris (WBSPI)~\cite{trokielewicz2018cross} contains both NIR images and RGB images of the same individuals. The lighting differences between the NIR and RGB acquisitions cause pupil size differences. For our experiments, we utilize the red channel from the RGB images to evaluate the robustness of {\it EyePreserve} when processing images outside the standard NIR spectrum.}

Finally, the \textbf{CASIA-Iris-Lamp (CIL)} test dataset~\cite{CASIA} consists of iris images taken using a handheld OKI iris sensor at different illumination levels.

\siamulnew{It is crucial to note that the evaluations on the Q-FIRE, CIL, and WBSPI datasets represent strict cross-dataset, zero-shot settings. The {\it EyePreserve} model was trained exclusively on WBPD and CSOSIPAD data and had never encountered the sensor characteristics or capture environments of Q-FIRE, CIL, or WBSPI during training.}

\subsection{Subject-Disjoint Train/Validation/Test Splits}

The train, validation, and test subsets are {\bf eye-disjoint}. These eye-disjoint splits are imperative to ensure that the model is learning generic, biology-based iris muscle movements rather than dynamics specific to the eyes of subjects represented in the training dataset. For reproducibility, all split definitions are offered as metadata along with this paper.

%% file: sections/4_iris_texture_deformation_model.tex
\section{Iris Texture Deformation Model}
\label{sec:methodology}

\begin{figure*}[!ht]
\centering
\includegraphics[width=\linewidth]{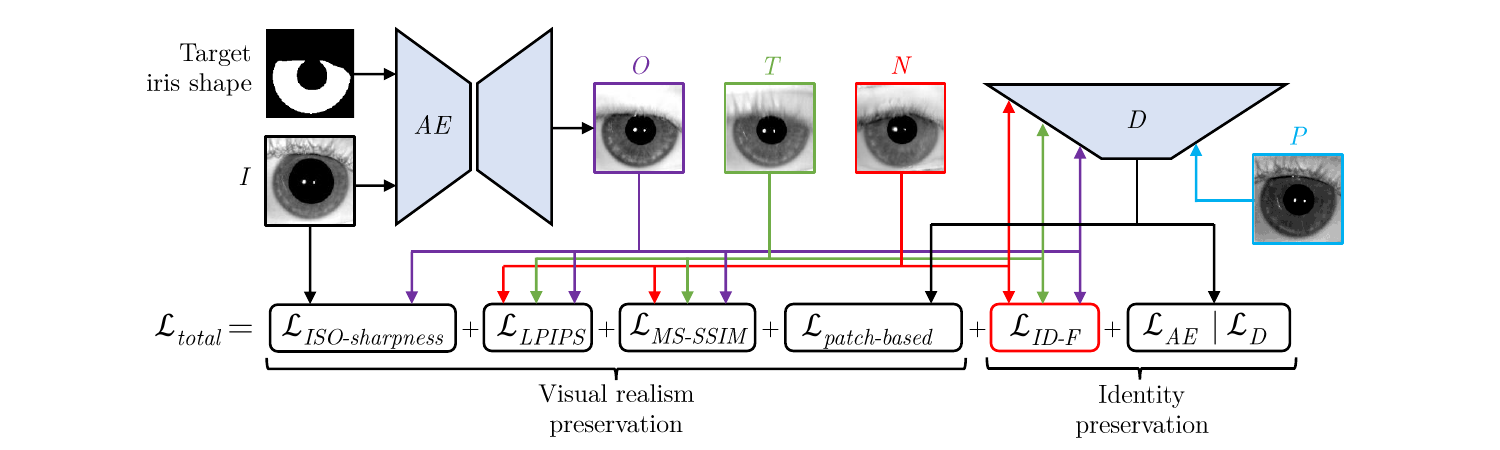}
\setlength{\belowcaptionskip}{-8pt}
\vspace{-12pt}
\caption{Illustration of the training mechanism with all loss function components explained in Sections \ref{sssec:identity-preservation} -- \ref{sssec:triplet}. Symbols: $I$ -- input image, $O$ -- output image, $T$ -- target sample, $N$ -- negative (impostor) sample, $P$ -- positive (genuine) sample, $D$ -- discriminator, $AE$ -- autoencoder}
\label{fig:train}
\end{figure*}

\siamulnew{To carry out identity-preserving iris texture deformation and generate realistic variations across different pupil sizes, we formulate this task as a mask-guided, image-to-image translation problem. Specifically, we train an autoencoder, guided by a novel triplet-adversarial objective and a filter-based identity loss operating over a fully differentiable pipeline, which takes an undeformed iris image and a target binary mask to synthesize the deformed output.} Fig.~\ref{fig:train} provides an overview of how this autoencoder is trained. 

\siamulnew{The selection of our composite loss function is motivated by the inherent trade-off between biometric utility and perceptual realism. Standard L1 or L2 losses typically yield blurry reconstructions by averaging high-frequency details. To counteract this, we utilize the LPIPS and PatchGAN adversarial losses to enforce high-frequency perceptual fidelity. However, because perceptual losses optimize for human vision rather than identity verification, we anchor the biometric identity using an Identity Filter-based loss, a novel Triplet-Adversarial formulation (to capture identity features beyond those specified by the fixed filters), and the ISO quality metric-based sharpness loss. This specific combination ensures that the model preserves the low-frequency anatomical structures critical for iris matchers like ArcIris and DGR, while generating textures that appear realistic to human examiners.}

Subsections below discuss the proposed training strategy and model specifics, grouped into four categories: identity preservation, image/iris realism preservation, formulation of the triplet-based loss function combining all required components, and model architecture.

\subsection{Identity Preservation}
\label{sssec:identity-preservation}

We utilize two loss components penalizing the model for not preserving the identity in a synthesized sample. The first, {\bf filter-based identity preservation} loss component $\mathcal{L}_{ID\text{-}\mathcal{F}}$ is defined as:
\begin{equation}
\mathcal{L}_{ID\text{-}\mathcal{F}}(I_1, I_2) = \|\,\mathcal{F}_{iris} \circledast N_D(I_1) - \mathcal{F}_{iris} \circledast N_D(I_2)\,\|
\end{equation}
where $I_1$ and $I_2$ are iris images being compared, $N_D$ denotes Daugman's iris image normalization~\cite{daugman1993high}, $\mathcal{F}_{iris}$ denotes a set of iris feature extraction filters, $\circledast$ denotes the convolution operation, and $\|\cdot\|$ denotes the $L_1$ norm. To create a comprehensive filter set assessing the biometric similarity between images, this loss component utilizes filters used in two open-source approaches: Gabor wavelets~\cite{othman2016osiris} and human perception-derived filters \cite{czajka2019domain}.

The second identity-preservation loss component is based on \textbf{adversarial training}. In generative modeling problems with classes of objects, the most commonly used architectural modification to the discriminator is the use of an auxiliary classifier~\cite{odena2017conditional} that assesses how well the discriminator synthesizes class exemplars. Thus, a naive approach would be to treat each identity as a separate class and train an auxiliary classification model along with the discriminator. However, this assumes that the total number of identities available in the training dataset is the complete set of identities possible.
%
Instead, our approach utilizes an adversarial identity loss, which allows the discriminator to learn embeddings from random triplets representing the same pupil-to-iris ratio and treat the output of the autoencoder as a separate identity, pushing its embedding away from the target (anchor) and impostor (negative) identities. On the other hand, the autoencoder learns to produce an image that has the same embedding (from the discriminator) as the target (anchor) image but is different from the impostor (negative) image. We can formally define this adversarial loss component as follows. First, let $\mathcal{L}_{cos}$ be the cosine loss, scaled to the $\langle0,1\rangle$ range:
\begin{equation}
\mathcal{L}_{cos}(V_1, V_2) = \big(1 - CosSim(V_1, V_2)\big)
\end{equation}

\noindent
where $CosSim$ is the cosine similarity. Let $I$ be the input (undeformed) image, $T$ be the target image, $O$ be the output (deformed) image, $P$ be the positive image that has the same identity as $I$ and pupil-to-iris ratio as $T$, and $N$ be the negative image that has a different identity but the same pupil-to-iris ratio as $T$. Let $AE$ denote the autoencoder and $D$ denote the discriminator such that $D(x)$ represents the embedding for an image $x$. For the loss of the autoencoder $\mathcal{L}_{AE}$, we have:

\begin{equation}
\begin{aligned}
\mathcal{L}_{AE} &= \mathcal{L}_{cos}\big(D(O), D(T)\big) + \mathcal{L}_{cos}\big(D(O), D(P)\big)\\
&+ \max\Big(\mathcal{L}_{cos}\big(D(O), D(I)\big) - margin_{IT},\,\,0\Big)\\
&+ \max\Big(margin_{NT} - \mathcal{L}_{cos}\big(D(O), D(N)\big),\,\,0\Big) 
\end{aligned}
\label{eqn:aeloss}
\end{equation}

\noindent
where
\begin{displaymath}
margin_{IT} = \mathcal{L}_{cos}\big(D(I), D(T)\big) 
\end{displaymath}
and
\begin{equation}
margin_{NT} = \mathcal{L}_{cos}\big(D(N), D(T)\big).
\label{eqn:marginNT}
\end{equation}

Thus, to train the autoencoder, we first minimize the distance of the output image to both the target and the positive images (first line in eq. \eqref{eqn:aeloss}). Secondly, we also minimize the distance between the output image and the input image as they belong to the same identity, but only as much as the loss between input and target, since the input image has a different pupil size than the target (second line in eq. \eqref{eqn:aeloss}). Finally, we maximize the loss between the output and negative images, but only as much as the loss between the target and negative images (last line in Eq.~\eqref{eqn:aeloss}).

For the loss of the discriminator $\mathcal{L}_D$, we have:
\begin{equation}
\begin{aligned}
\mathcal{L}_D &= \max\Big(\mathcal{L}_{cos}\big(D(T), D(P)\big) - \mathcal{L}_{cos}\big(D(T), D(N)\big) \\
&+ margin_D,\,\,0\Big) \\
&+ \max\Big(margin_{NTP} - \mathcal{L}_{cos}\big(D(O), D(T)\big),\,\,0\Big) \\
&+ \max\Big(margin_{NTP} - \mathcal{L}_{cos}\big(D(O), D(P)\big),\,\,0\Big) \\
&+ \max\Big(margin_{NTP} - \mathcal{L}_{cos}\big(D(O), D(N)\big),\,\,0\Big),
\end{aligned}
\end{equation}
where $margin_D$ is a hyperparameter that defines the minimum separation required between the anchor and the positive images, while also specifying the maximum separation that has to be retained between the anchor and the negative images, and $margin_{NTP}$ is an estimate of the maximum possible loss between impostor pairs of images with the same pupil size, and is defined as:
\begin{displaymath}
margin_{NTP} = max(margin_{NT}, margin_{NP})
\end{displaymath}
where $margin_{NP} = \mathcal{L}_{cos}\big(D(N), D(P)\big)$, and $margin_{NT}$ is defined by \eqref{eqn:marginNT}. In our training, we found that a $margin_D$ of 0.3 works best.

Therefore, the first part of the loss $\mathcal{L}_D$ is a classical triplet margin loss between the target, positive, and negative images. This allows the discriminator to learn to distinguish iris images of different identities. In addition, with the subsequent components of the loss $\mathcal{L}_D$, the discriminator learns to distinguish the autoencoder-generated image as different from the target, positive, and negative images.
In summary, the autoencoder tries to minimize the distance between the embeddings of its output image and the original (target) image, while the discriminator tries to maximize this distance while simultaneously learning to distinguish between different identities.

\subsection{Realism Preservation}
\label{sssec:realism-preservation}
\noindent
To induce visual realism in the output images, we use a combination of the following losses.

\vskip2mm\noindent \textbf{Perceptual Losses:} We utilize two popular perceptual losses: the Learned Perceptual Image Patch Similarity (LPIPS) loss~\cite{zhang2018perceptual} $\mathcal{L}_{LPIPS}$, and Multi-Scale Structural Similarity (MS-SSIM) based loss~\cite{wang2003multiscale} $\mathcal{L}_{MS-SSIM}$.

\vskip2mm\noindent
\textbf{Adversarial Patch Loss:} In our discriminator, we utilize the PatchGAN architecture~\cite{isola2017image} that
learns to distinguish each image patch as real or fake. This image patch-based loss
enables the model to enhance higher spatial frequencies  
to improve human visual perception of the synthesized images.

\vskip2mm\noindent
\textbf{ISO Sharpness Loss:} ISO/IEC 29794-6 \verb+SHARPNESS+ metric utilizes a single filtering kernel engineered to capture frequencies in an iris image that are essential for iris recognition. Let $ISO$ be the ISO \verb+SHARPNESS+ score, $O$ be the output image, and $In$ be the input image. Then the ISO-sharpness loss
\begin{equation}
\mathcal{L}_{sharp}(O, In) = \max\big(ISO(In) - ISO(O),\,\,0\big)
\end{equation}
This loss ensures that the output image has the same or higher ISO \verb+SHARPNESS+ score as the input image provided to the autoencoder. 

\subsection{Triplet Formulation}
\label{sssec:triplet}

A direct comparison between the output and target images should encourage a model to focus more on features that are common across the images. Thus, especially for losses such as $\mathcal{L}_1$ and $\mathcal{L}_2$, this method of direct image comparison fails to capture distinctive features or to encourage high-frequency crispness. Consequently, for all of our identity-preservation losses ($\mathcal{L}_{ID\text{-}\mathcal{F}}$, $\mathcal{L}_{AE}$, $\mathcal{L}_D$) and realism-preservation losses ($\mathcal{L}_{LPIPS}$, $\mathcal{L}_{MS-SSIM}$, $\mathcal{L}_{sharp}$) we employ a triplet-based formulation: 

\begin{equation}
\mathcal{L} \vcentcolon= \mathcal{L}(O, T) + \max\big(margin - \mathcal{L}(O, N),\,\,0\big)
\end{equation}
where
\begin{equation}   
margin = \mathcal{L}(T, N).
\end{equation}
and $O$ is the output image, $T$ is the target image, and $N$ is the negative (impostor) image.

The margin in the usual triplet loss defines the minimum possible value by which the anchor-negative loss should be higher than the anchor-positive loss. However, the margin in our formulation directly defines a value for the minimum possible anchor-negative loss. That is, in our triplet formulation, we utilize the fact that the output image should be as different from the impostor image as the target image should be, and take the loss between the target and the impostor image as the margin.
By moving the output image away from an impostor image based on how much the target image is different from the impostor, while simultaneously moving the output image closer to the target, we encourage the model to generate images with more distinctive features rather than generating a blurry image capturing the common features across images.

\siamulnewnew{\subsection{Composite Loss Formulation}
\label{sssec:composite-loss}

To train the autoencoder, the individual identity and realism preservation components described in Sections \ref{sssec:identity-preservation} and \ref{sssec:realism-preservation} are combined into a single total loss:

\begin{equation}
\begin{aligned}
\mathcal{L}_{total} &= \mathcal{L}_{ID\text{-}\mathcal{F}} + \mathcal{L}_{AE} + \mathcal{L}_{LPIPS} \\
&\quad + \mathcal{L}_{MS-SSIM} + \mathcal{L}_{sharp} + \mathcal{L}_{patch\text{-}based}
\end{aligned}
\end{equation}

As shown in the above equation and in Fig.~\ref{fig:train}, all loss components are applied with equal weights. While it is common to manually tune weights to balance competing objectives, our empirical observations during early model development showed that introducing specific scalar weights did not yield significant performance improvements over this unweighted sum.}

\subsection{Autoencoder Architecture}
This work utilizes an attention-based nested U-Net \cite{li2020anu, li2020attention} with the downsampling operation changed from max-pooling to bilinear downsampling. \siamulnew{Our training method is independent of the specific network architecture used for the autoencoder. While we demonstrate its efficacy using an attention-based nested U-Net, which was chosen for its strong performance in fine-grained texture modeling, the modular design ensures that emerging autoencoder architectures with advanced generation capabilities can be seamlessly integrated in the future without altering the overarching framework.}


\subsection{Target Mask Estimation for a particular Pupil-to-Iris ratio}
\begin{figure}[!ht]
\centering
\includegraphics[width=\linewidth]{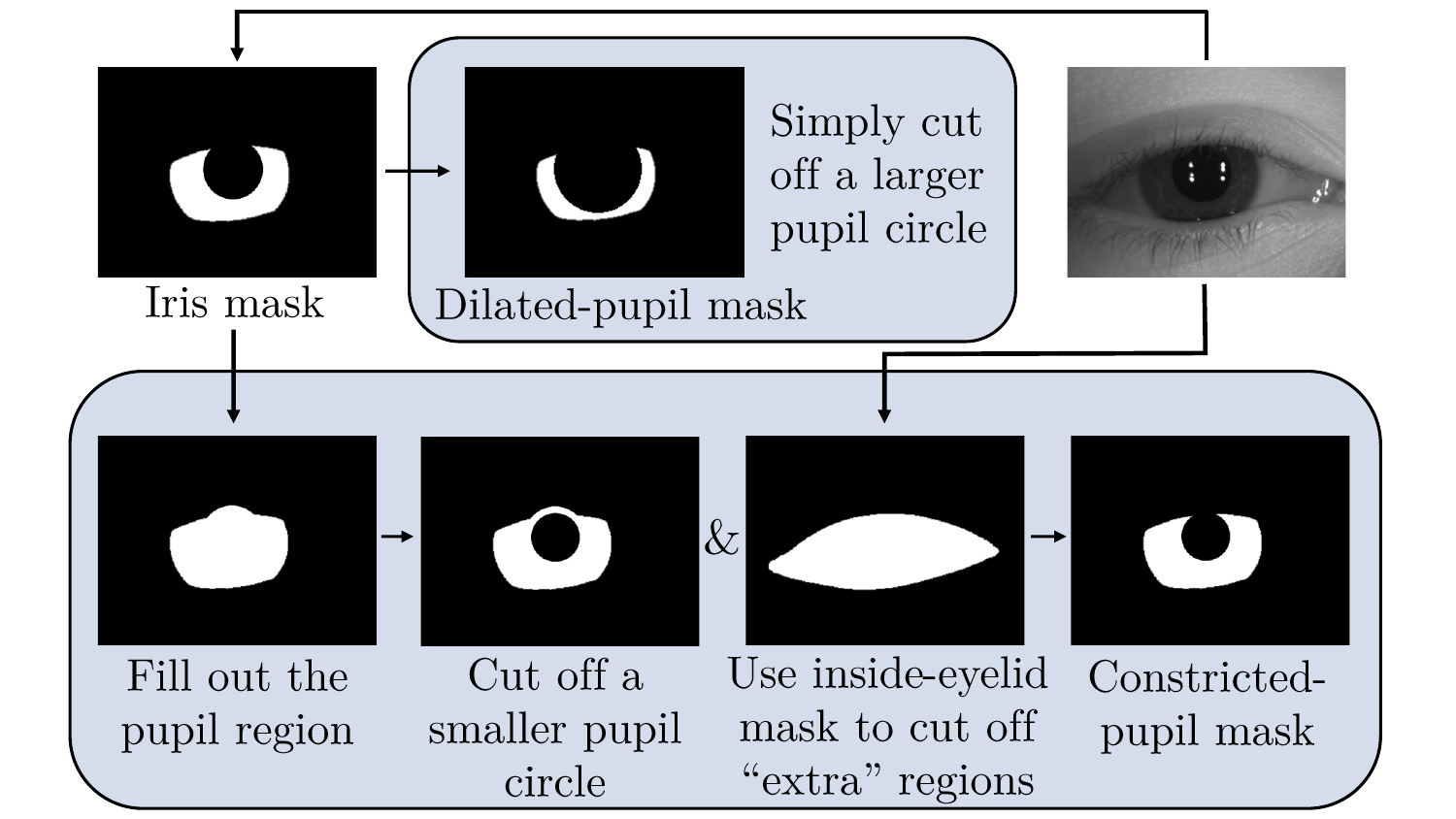}
\setlength{\belowcaptionskip}{-7pt}
\vspace{-7pt}
\caption{Illustration of the target iris image mask estimation. {\bf Upper row:} Getting the target iris mask for dilated pupils is easy, and cutting a larger pupil circle in the original mask. {\bf Bottom row:} For the constricted-pupil iris mask, we need to fill out the pupil region and then cut a smaller pupil circle. However, when the pupil is partially occluded by eyelids, there are residual regions that we remove by using an ``inside-eyelid'' mask.}
\label{fig:mask_estimation}
\end{figure}

The identity preservation component of the proposed {\it EyePreserve} method requires comparing images of synthetically-generated images of irises having different pupil sizes and different palpebral fissure (eyelid opening). In such a comparison, we can deform one of the images to match the pupil size of the other image using the segmentation mask of the latter using the {\it EyePreserve} model. If we simply want to constrict/dilate an iris image given a new pupil ``size,'' we can detect approximate circles for the pupil and iris using existing models, \eg the one proposed in \cite{vance2022deception}, and generate a circular mask corresponding to a new pupil radius.
However, this approach would lead to hallucinated iris textures if the input iris image is occluded by eyelids. Thus, for synthesizing new iris images with a dilated pupil, we cut a larger pupil circle from the original (input) mask. 
To synthesize a new iris image with a constricted pupil, we fill out the pupil circle in the original (input) mask and then cut out a smaller pupil circle from it. This still leaves residual regions in the mask (from the larger pupil that was filled out) if there is occlusion by the eyelids. So, we use a segmentation model that segments the region inside the eyelids to discard any part of the mask outside the eyelid boundary. 
Figure~\ref{fig:mask_estimation} illustrates this procedure. 

\vspace{-3pt}

\subsection{Synthesis of Irises from Non-existing Identities} 

\siamulnew{To offer the implementation of the entire framework for pupil size-varying, identity-preserving iris image synthesis, which also includes image synthesis for non-existing identities, we include StyleGAN3-based~\cite{karras2021alias} and Denoising Diffusion Probabilistic Model (DDPM)-based~\cite{ho2020denoising} iris image generators. Both models were trained on a combination of the WBPD and CSOSIPAD datasets. While these networks generate realistic irises, their resulting pupil sizes are random and not directly controllable. Thus, we can synthesize an iris image of an arbitrary pupil size, and then utilize the {\it EyePreserve} model to vary that pupil size. Any state-of-the-art iris image generation model can certainly be added without any changes to the existing {\it EyePreserve} framework.}

%% file: sections/5_results.tex
\section{Experiments and Results}
\label{sec:results}

\subsection{Iris Recognition Methods}
To evaluate the identity-preserving capabilities of our system for both real and synthetically-generated iris images, we carry out evaluations 
\siamul{ 
with three iris recognition methods: i) a neural network trained for iris recognition using the ArcFace loss (ArcIris)~\cite{khan2026lowering}, which is the first open-source iris recognition method included in the NIST IREX X leaderboard~\cite{IREX_X_URL}, ii) Dynamic Graph Representation for Occlusion Handling in Biometrics (DGR)~\cite{ren2020dynamic}, and iii) WorldCoin-IRIS (WCI)~\cite{wldiris}}. 

\siamul{Using square {\it EyePreserve}-generated images created incorrect results in the case of the WCI matcher. \siamulnew{Furthermore, attempting to pad or crop and then resize these square images to an ISO-standard resolution failed to resolve the issue. The matcher produced inconsistent results that were highly sensitive to the amount of padding or cropping applied, even though the resulting images complied with the ISO/IEC 29794-6 recommendations.}
Thus, instead of using the entire WCI iris recognition pipeline, we only use the WCI Gabor filters, while the segmentation method is the same across all three matchers used, which allows for more comparable and interpretable results.}

\subsection{Uncertainty Estimation}
To estimate the uncertainty of the obtained results, and to report interval error estimates to assess statistical significance of the observed differences between methods,  
we randomly sample 10\% of genuine and impostor comparison scores 1,000 times (with replacement) when calculating each Receiver Operating Characteristic (ROC) curve. We then report the average Area Under the ROC Curve (AUC) and average decidability ($d'$) along with standard deviations of the AUC calculated over these 1,000 samples. \siamulnew{The ROC curves for the full set of scores for each method and dataset combination is provided in the supplementary materials.}

\subsection{Identity Preservation Assessment}
\label{sec:identity_preservation}

\begin{figure}[!ht]
\centering
\includegraphics[width=\linewidth]{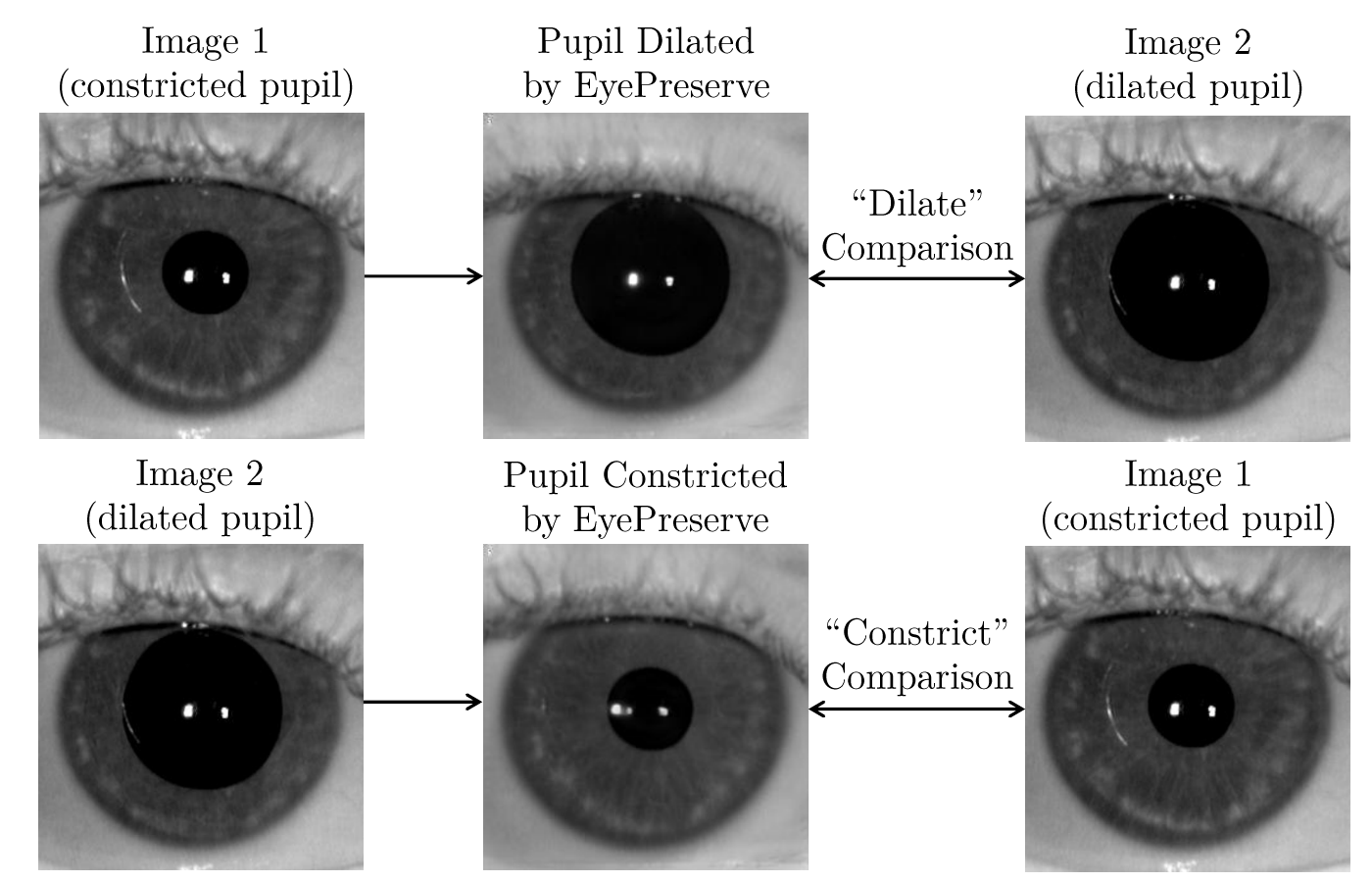}
\setlength{\belowcaptionskip}{-7pt}
\caption{\siamul{
The {\it EyePreserve} model deforms an iris image to either dilate a small pupil to align with the larger one, or constrict a large pupil to align with the smaller one. This gives two comparison scores for each pair of images being matched.  
We evaluated minimum, maximum, and mean fusion for the scores from these comparison modes, and found that mean performed the best.}}
\label{fig:dilate_constrict_comp}
\end{figure}

\siamul{The {\it EyePreserve} model can both constrict and dilate the pupil. This gives two options when two iris images with different pupil sizes are compared: i) dilate the smaller-pupil iris image and compare with the original (larger-pupil) image, or ii) constrict the larger-pupil iris image and compare with the original (smaller-pupil) image, as shown in  Fig.~\ref{fig:dilate_constrict_comp}. In the experiments on validation data, we found that taking the mean of the two scores provides the best results, which are reported in the following sections.}

A simple measure for the degree of pupil size change is the difference in the pupil-to-iris ratio $\Delta = \| p_1/i_1 - p_2/i_2\|$, where $p$ and $i$ are pupil and iris radii, respectively. $\Delta$ is large if one eye has a significantly constricted pupil and the other has a significantly dilated pupil, and small if pupil sizes are similar. 

The following subsections demonstrate the identity preservation capabilities of the proposed model when applied to slightly (small $\Delta$), moderately, or significantly (large $\Delta$) deformed iris texture. We show that {\it EyePreserve} does not degrade the matching performance when the correction for pupil size is small, and it does improve the performance significantly when it is applied to correct large differences in pupil size.

\subsubsection{Performance for Extreme Iris Deformations ($\Delta\geq0.25$)}
First, we analyze how  {\it EyePreserve}  performs when the difference in pupil size between the probe and gallery images is significant. We utilize the \siamul{WBPD-plus} dataset, which was specifically collected to study iris texture deformation under large pupil size variations, and thus contains a wide range of pupil sizes for the same eyes. 

\begin{figure*}[!htbp]
\centering
\begin{subfigure}{0.3\textwidth}
    \centering
    \includegraphics[width=\linewidth]{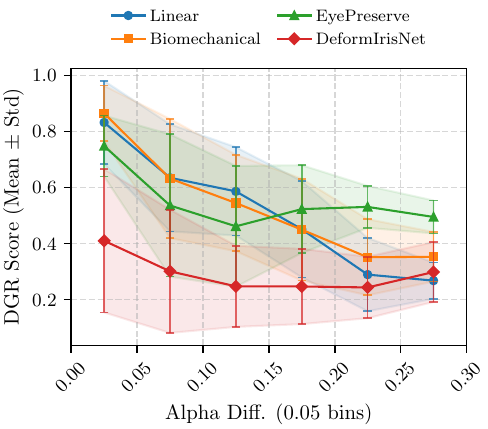}
    \caption{}
\end{subfigure} 
\hfill
\begin{subfigure}{0.3\textwidth}
    \centering
    \includegraphics[width=\linewidth]{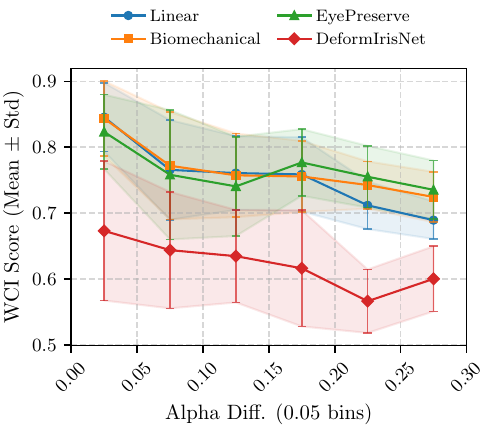}
    \caption{}
\end{subfigure}
\hfill
\begin{subfigure}{0.3\textwidth}
    \centering
    \includegraphics[width=\linewidth]{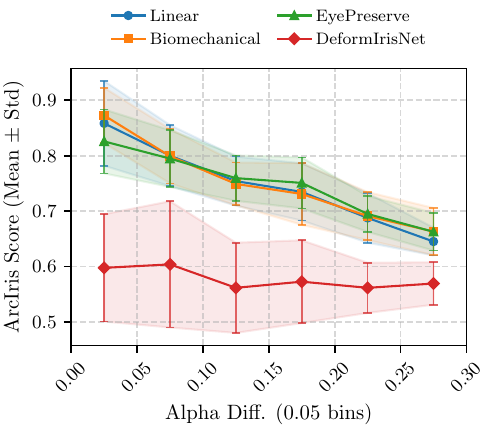}
    \caption{}
\end{subfigure}
\setlength{\belowcaptionskip}{-7pt}
\caption{The similarity scores offered by (a) DGR \cite{ren2020dynamic}, (b) WCI~\cite{wldiris}, and (c) ArcIris~\cite{khan2026lowering} as a function of the pupil-to-iris ratio difference ($\Delta$) between the probe and gallery samples for WBPD+ test set.\vspace{10pt}}
\label{fig:wbpdfwise}
\end{figure*}

\siamul{For the WBPD-plus dataset, we provide a granular analysis of AUC and $d'$ results across varying levels of $\Delta$, Tab.~\ref{tab:wbpdresults}. Additionally, Figure\ref{fig:wbpdfwise} illustrates the relationship between mean genuine comparison scores and $\Delta$, highlighting how the comparison score changes with the degree of iris texture deformation. Due to the high fidelity of the WBPD data, a low number of identities in the test set, and the data curation that was performed to remove almost-closed eyes, ArcIris achieved almost perfect AUC on the WBPD-plus test set with any method of iris deformation. As such, we omit the AUCs obtained for ArcIris in Tab.~\ref{tab:wbpdresults}.}

\siamul{As shown in Tab.~\ref{tab:wbpdresults}, traditional linear normalization and biomechanical models begin to fail as $\Delta$ exceeds 0.2. In the most extreme cases ($\Delta>0.25$), {\it EyePreserve} significantly outperforms the baselines, maintaining an AUC of 0.997 (DGR) and 0.995 (WCI), whereas for the linear model AUC=0.982 and 0.994, respectively. This trend is further illustrated by the mean similarity scores in Fig.~\ref{fig:wbpdfwise}. 

For small $\Delta$, {\it EyePreserve} yields slightly lower performance due to the autoencoder's reconstruction artifacts outweighing the need for minimal iris texture deformations. However, as $\Delta$ increases, non-linear iris texture deformations start to dominate. In these cases, the superior modeling capability of {\it EyePreserve} becomes evident, resulting in higher performance compared to other methods.}

\begin{table*}[ht]
\centering
\tabcolsep 3pt
\caption{\label{tab:wbpdresults}AUCs, $d'$, and EER averaged over 1,000 re-sampled (with replacement) sets of 10\% of all cross-comparison scores on the WBPD-plus test set for different bins of $\Delta$ (difference in pupil-to-iris ratio between images being compared). Due to the high fidelity of the WBPD+ data, a low number of identities in the test set (only 8 unique eyes), and the data curation that was performed to remove almost-closed eyes, ArcIris achieves almost perfect AUC on the test set with any method of iris deformation (except DeformIrisNet), so we only show $d'$ for the ArcIris matcher.}
\resizebox{\textwidth}{!}{%
\begin{tabular}{llcccccccc}
\toprule
\multirow{2}{*}{$\Delta$} & \multirow{2}{*}{Methods}  & \multicolumn{2}{c}{ArcIris} & \multicolumn{3}{c}{DGR} & \multicolumn{3}{c}{WCI} \\
\cmidrule(lr){3-4} \cmidrule(lr){5-7} \cmidrule(lr){8-10}
 & & $d'$ & EER($\%$) & AUC & $d'$ & EER($\%$) & AUC & $d'$ & EER($\%$) \\
\cline{1-10}
\multirow{4}{*}{$[0, 0.05)$} & Linear & $5.13\pm0.01$ & $0.153\pm0.016$ & \bm{$0.996\pm0.000$} & $3.62\pm0.01$ &  $2.322\pm0.046$ &  \bm{$0.999\pm0.000$} & $3.60\pm0.06$ & $0.697\pm0.028$ \\ \cline{2-10}
 & Biomech & $5.12\pm0.02$ & \bm{$0.149\pm0.012$} & $0.995\pm0.000$ & $3.64\pm0.01$  & $2.323\pm0.051$  & \bm{$0.999\pm0.000$} &  $3.60\pm0.04$ &  $0.698\pm0.028$ \\ \cline{2-10}
 & DeformIrisNet & $1.68\pm0.01$ & $23.380\pm0.144$ & $0.947\pm0.001$ & $2.08\pm0.01$ & $13.765\pm0.108$ &  $0.895\pm0.001$ & $1.65\pm0.01$ & $20.122\pm0.130$ \\ \cline{2-10}
 & EyePreserve & \bm{$5.33\pm0.01$} & $0.190\pm0.019$ & \bm{$0.996\pm0.000$} & \bm{$4.53\pm0.01$} & \bm{$1.096\pm0.034$} & \bm{$0.999\pm0.000$} & \bm{$5.11\pm0.11$} & \bm{$0.379\pm0.020$} \\
\cline{1-10}
\multirow{4}{*}{$[0.05, 0.1)$} & Linear & $6.00\pm0.02$ & \bm{$0.318\pm0.022$} & $0.994\pm0.000$ & $3.57\pm0.01$ & $2.801\pm0.054$ & \bm{$0.998\pm0.000$} & $3.15\pm0.08$ & $0.473\pm0.021$
 \\ \cline{2-10}
 & Biomech & $5.99\pm0.03$ & $0.319\pm0.020$ & $0.995\pm0.000$ & $3.42\pm0.01$ & $2.800\pm0.055$ & \bm{$0.998\pm0.000$} & $3.12\pm0.08$ & $0.469\pm0.021$\\ \cline{2-10}
 & DeformIrisNet & $1.53\pm0.01$ & $26.215\pm0.149$ & $0.937\pm0.001$ & $1.96\pm0.01$ & $16.351\pm0.120$ & $0.869\pm0.001$ & $1.44\pm0.02$ & $24.049\pm0.137$ \\ \cline{2-10}
 & EyePreserve & \bm{$6.07\pm0.02$} & $0.404\pm0.027$ & \bm{$0.996\pm0.000$} & \bm{$4.67\pm0.02$} & \bm{$1.024\pm0.038$} & \bm{$0.998\pm0.000$} & \bm{$4.86\pm0.13$} & \bm{$0.331\pm0.018$} \\
\cline{1-10}
\multirow{4}{*}{$[0.1, 0.15)$} & Linear & \bm{$6.50\pm0.02$} & $0.143\pm0.015$ & $0.995\pm0.000$ & $3.65\pm0.01$ & $2.340\pm0.047$ & $0.998\pm0.000$ & $3.85\pm0.12$ & $0.294\pm0.019$ \\ \cline{2-10}
 & Biomech & $6.49\pm0.01$ & \bm{$0.140\pm0.011$} & $0.994\pm0.000$ & $3.42\pm0.01$ & $2.320\pm0.061$ & \bm{$0.998\pm0.000$} & $3.85\pm0.13$ & $0.297\pm0.019$\\ \cline{2-10}
 & DeformIrisNet & $1.54\pm0.01$ & $25.940\pm0.154$ & $0.933\pm0.001$ & $1.96\pm0.01$ &  $17.005\pm0.117$ & $0.862\pm0.001$ & $1.46\pm0.01$ & $25.269\pm0.140$ \\ \cline{2-10}
 & EyePreserve & $6.30\pm0.02$ & $0.183\pm0.018$ & \bm{$0.998\pm0.000$} & \bm{$5.34\pm0.02$} & \bm{$0.496\pm0.027$} & \bm{$0.998\pm0.000$} & \bm{$5.55\pm0.12$} & \bm{$0.208\pm0.015$} \\
\cline{1-10}
\multirow{4}{*}{$[0.15, 0.2)$} & Linear & \bm{$6.27\pm0.02$} & \bm{$0.030\pm0.007$} & $0.994\pm0.000$ & $3.58\pm0.01$ & $2.857\pm0.053$ & \bm{$0.999\pm0.000$} & $4.31\pm0.16$ & $0.329\pm0.019$ \\ \cline{2-10}
 & Biomech & \bm{$6.27\pm0.03$} & \bm{$0.030\pm0.006$} & $0.994\pm0.000$ & $3.43\pm0.01$ & $2.837\pm0.073$ & \bm{$0.999\pm0.000$} & $4.19\pm0.16$ & $0.322\pm0.022$ \\ \cline{2-10}
 & DeformIrisNet & $1.49\pm0.01$ & $24.289\pm0.147$ & $0.927\pm0.001$ & $1.99\pm0.01$ & $15.898\pm0.118$ & $0.849\pm0.001$ & $1.42\pm0.01$ & $28.108\pm0.148$ \\ \cline{2-10}
 & EyePreserve & $5.98\pm0.02$ & $0.071\pm0.010$ & \bm{$0.999\pm0.000$} & \bm{$5.33\pm0.02$} & \bm{$0.778\pm0.030$} & \bm{$0.999\pm0.000$} & \bm{$5.80\pm0.10$} & \bm{$0.293\pm0.017$} \\
\cline{1-10}
\multirow{4}{*}{$[0.2, 0.25)$} & Linear & $5.32\pm0.01$ & $0.076\pm0.011$ & $0.992\pm0.000$ & $3.01\pm0.01$ & $3.558\pm0.061$ & $0.995\pm0.000$ & $2.78\pm0.09$ & $0.740\pm0.028$ \\ \cline{2-10}
 & Biomech & $5.33\pm0.01$ & $0.072\pm0.014$ & $0.992\pm0.000$ & $3.45\pm0.01$ & $3.548\pm0.061$  & $0.995\pm0.000$ & $2.82\pm0.09$ & $0.730\pm0.020$ \\ \cline{2-10}
 & DeformIrisNet & $1.42\pm0.01$ & $25.536\pm0.145$  & $0.901\pm0.001$ & $1.80\pm0.01$ &  $18.113\pm0.122$ & $0.878\pm0.001$ & $1.33\pm0.01$ & $23.547\pm0.135$ \\ \cline{2-10}
 & EyePreserve & \bm{$5.83\pm0.01$} & \bm{$0.056\pm0.008$} & \bm{$0.997\pm0.000$} & \bm{$5.14\pm0.01$} & \bm{$1.151\pm0.033$} & \bm{$0.997\pm0.000$} & \bm{$4.50\pm0.10$} & \bm{$0.678\pm0.028$} \\
\cline{1-10}
\multirow{4}{*}{$(0.25, \infty)$} & Linear & $4.89\pm0.01$ & $0.348\pm0.018$ & $0.982\pm0.000$ & $3.16\pm0.01$ & $5.785\pm0.074$ &  $0.994\pm0.000$ & $2.17\pm0.07$ & $1.089\pm0.035$ \\ \cline{2-10}
 & Biomech & $4.89\pm0.01$ & $0.318\pm0.022$ & $0.985\pm0.000$ & $2.81\pm0.01$ & $5.745\pm0.074$ & $0.995\pm0.000$ & $2.18\pm0.07$ & $1.070\pm0.035$ \\ \cline{2-10}
 & DeformIrisNet & $1.41\pm0.01$ & $22.040\pm0.138$ & $0.943\pm0.001$ & $2.07\pm0.01$ & $15.588\pm0.115$ & $0.873\pm0.001$ & $1.24\pm0.02$ & $22.707\pm0.128$ \\ \cline{2-10}
 & EyePreserve & \bm{$5.24\pm0.01$} & \bm{$0.267\pm0.019$}  & \bm{$0.997\pm0.000$} & \bm{$4.14\pm0.01$} &  \bm{$3.143\pm0.058$} & \bm{$0.996\pm0.000$} & \bm{$3.32\pm0.08$} & \bm{$0.999\pm0.033$} \\
\bottomrule
\end{tabular}%
}
\end{table*}

\subsubsection{Performance in Realistic Conditions ($\Delta \geq 0.2$)} 

To analyze how our model performs with \siamul{typical} pupil size variations, we utilize the HWS dataset (see Sec. \ref{sec:testdata}). The pupil variation was achieved in this dataset by turning the ambient lighting on and off during the collection, which means that the iris muscles were not taken to their full range of motion. As such, the range of $\Delta$ values 
corresponds to real pupil size variations that may be observed in normal use of iris sensors more closely than the WBPD dataset.

We select sample pairs from the HWS dataset with the largest possible $\Delta$. To secure a sufficient number of samples, this translates to $\Delta \geq 0.2$, since the HWS dataset has very few pairs with $\Delta > 0.3$. Results presented in Tab. \ref{tab:allresults} suggest that the proposed {\it EyePreserve} approach outperforms all existing iris texture deformation models. Additionally, we can see that the only existing nonlinear deep learning-based iris deformation model, DeformIrisNet, does not generalize well to a new dataset. 

\subsubsection{Reconstruction Overhead at Low $\Delta$ ($\Delta \leq 0.1$)}

The goal of this third experimental scenario is to check whether there is a significant impact on the iris recognition performance when there is little or no change in pupil size. 
For this purpose, we consider three subsets: (a) the subset of the HWS set with $\Delta\leq0.1$, (b) the Q-FIRE dataset, and (c) the CIL dataset. While Q-FIRE and CIL datasets contain images with varying pupil sizes, the possible number of pairs with $\Delta\ge0.2$ would not secure statistically significant evaluations.

\siamul{According to the results in Table~\ref{tab:allresults}, the differences in iris recognition performance when a linear model and the {\it EyePreserve} model are applied are not statistically significant when the pupil difference is low. However, in most cases, the {\it EyePreserve} model seems to improve inter-class separability as observed from the higher decidability scores ($d'$). We hypothesize that this improvement is primarily caused by the elimination of pupil-size variation during the comparison process. By deforming both irises to a consistent pupil state (either through dilation or constriction as illustrated in Fig.~\ref{fig:dilate_constrict_comp}), {\it EyePreserve} mitigates the geometric noise associated with varying pupil sizes.

The experiments and results in this subsection confirm that the cost of propagating an iris image through the {\it EyePreserve} model, in a situation where there is no significant iris texture deformation needed, is negligible. This allows us to conclude that this nonlinear model improves iris recognition when the pupil size difference is large, and does not hurt the performance when there is no need for pupil size alignment.}

\subsubsection{\siamulnew{RGB-NIR Cross-Spectral Matching Performance}}
\siamulnew{
The objectives of this fourth experimental scenario are to evaluate the performance of the {\it EyePreserve} approach when applied to (a) visible-light (RGB) iris images, and (b) the cross-spectral (RGB vs NIR) matching. For this purpose, we utilize the Warsaw-BioBase-Smartphone-Iris (WBSPI) dataset. 

For our experiment, we extract and utilize only the red channel from the RGB images, which serves as the visible-light proxy for the matching process. Also, we strictly isolate the cross-spectral challenge by generating comparison pairs exclusively between RGB and NIR images. No intra-spectral pairs (i.e., RGB-to-RGB or NIR-to-NIR) are included in this evaluation subset. This pairing strategy is used because the differing illumination conditions required for RGB versus NIR captures naturally induce significant variations in pupil size.

The results for the WBSPI dataset in Table~\ref{tab:allresults} highlight two major takeaways regarding cross-spectral matching. First, baseline deep learning models struggle with the spectral gap. Extractors like ArcIris and DGR perform worse (especially apparent in the case of DGR) because they were trained almost exclusively on NIR data. In contrast, WCI, which relies on classical Daugman Gabor filters, proves much more robust. 

Second, the proposed {\it EyePreserve} model consistently delivers superior or highly competitive results. For ArcIris and DGR, it achieves the highest AUCs (0.950 and 0.859) and decidability scores ($d'$ of 2.33 and 1.62). Additionally, {\it EyePreserve} records the lowest EER for ArcIris (11.309), while its EER for DGR closely trails the top value achieved by the biomechanical model (22.659 vs. 22.302). When paired with WCI, {\it EyePreserve} nearly doubles the $d'$ score to 2.19 (compared to 1.12 for both the linear and biomechanical models) while maintaining a strong AUC of 0.966 (vs. 0.970) and an EER of 5.378 (vs. 5.103). Overall, even in the few metrics where {\it EyePreserve} does not achieve the absolute best value, its performance remains highly competitive. This demonstrates that although {\it EyePreserve} is trained solely on NIR images, it successfully learns pupil deformations that generalize well to the red channel of RGB images.

}

\begin{table*}[!htbp]
\vspace{15pt}
\centering
\caption{\label{tab:allresults}\siamul{AUC, $d'$, and EER obtained for ArcIris, DGR, and WCI matchers on HWS, QFIRE, CIL, and WBSPI datasets. Values are averaged over 1,000 bootstrap samples (10\% of all cross-comparisons sampled with replacement). The best-performing mean fusion results (cf. Fig.~\ref{fig:dilate_constrict_comp}) are reported.}}
\resizebox{\textwidth}{!}{%
\begin{tabular}{clccccccccc}
\toprule
\multirow{3}{*}{} & & \multicolumn{9}{c}{Matcher} \\ \cmidrule(lr){3-11}
 & & \multicolumn{3}{c}{ArcIris} & \multicolumn{3}{c}{DGR} & \multicolumn{3}{c}{WCI} \\ \cmidrule(lr){3-5} \cmidrule(lr){6-8} \cmidrule(lr){9-11}
Dataset & Method & AUC & $d'$ & EER($\%$) & AUC & $d'$ & EER($\%$) & AUC & $d'$ & EER($\%$) \\ \midrule
\multirow{4}{*}{\begin{tabular}[c]{@{}c@{}}HWS \\ ($\Delta\leq0.1$)\end{tabular}} & Linear & \bm{$0.996\pm0.001$} & $5.11\pm0.33$ & $0.669\pm0.071$ & $\bm{0.993\pm0.000}$ & $2.86\pm0.02$ & $3.911\pm0.130$ & $0.992\pm0.001$ & $5.01\pm0.04$ & $1.110\pm0.076$ \\ \cline{2-11}
 & Biomech & \bm{$0.996\pm0.001$} & $5.01\pm0.04$ & $0.668\pm0.069$ & $\bm{0.993\pm0.000}$ & $2.83\pm0.02$ & $3.664\pm0.119$ & $\bm{0.993\pm0.001}$ & $4.78\pm0.04$ & \bm{$1.050\pm0.084$} \\ \cline{2-11}
 & DeformIrisNet & $0.723\pm0.003$ & $0.93\pm0.01$ & $35.183\pm0.310$ & $0.701\pm0.003$ & $0.79\pm0.01$ & $36.462\pm0.316$ & $0.714\pm0.003$ &  $0.92\pm0.01$ & $35.800\pm0.314$ \\ \cline{2-11}
 & EyePreserve & \bm{$0.996\pm0.001$} & \bm{$5.23\pm0.04$} & \bm{$0.639\pm0.064$} & $0.991\pm0.000$ & $\bm{3.65\pm0.03}$  & \bm{$1.694\pm0.084$} & $0.991\pm0.001$ &   $\bm{5.50\pm0.06}$ & $1.284\pm0.085$ \\ \midrule
\multirow{4}{*}{\begin{tabular}[c]{@{}c@{}}HWS \\ ($\Delta\geq0.2$)\end{tabular}} & Linear & $0.997\pm0.000$ & \bm{$8.26\pm0.33$} & $0.905\pm0.225$ & $0.981\pm0.002$ & $2.76\pm0.05$ & $9.053\pm0.440$ & $0.980\pm0.003$ & $4.20\pm0.12$ &  $2.802\pm0.348$ \\ \cline{2-11}
 & Biomech & $0.997\pm0.000$ & \bm{$8.26\pm0.32$} & $0.893\pm0.212$ & $0.981\pm0.002$ & $2.81\pm0.05$ & $6.514\pm0.404$ & $0.979\pm0.003$ & $3.93\pm0.09$ & $2.966\pm0.281$ \\ \cline{2-11}
 & DeformIrisNet & $0.653\pm0.008$ & $0.62\pm0.03$ & $39.493\pm0.629$ & $0.723\pm0.006$ & $0.70\pm0.02$ & $34.069\pm0.591$ & $0.627\pm0.008$ & $0.46\pm0.03$ & $40.999\pm0.672$\\ \cline{2-11}
 & EyePreserve & \bm{$0.999\pm0.000$}  & $7.17\pm0.15$ & \bm{$0.682\pm0.164$} & $\bm{0.989\pm0.001}$ & $\bm{4.02\pm0.08}$ & \bm{$4.111\pm0.284$} & $\bm{0.984\pm0.001}$ & $\bm{4.75\pm0.16}$ & \bm{$2.649\pm0.366$} \\ \midrule
\multirow{4}{*}{QFIRE} & Linear & $0.996\pm0.000$ & $5.42\pm0.02$ & $1.180\pm0.031$ & $0.974\pm0.000$ & $2.81\pm0.01$ & $8.648\pm0.064$ & $0.992\pm0.000$ & $2.94\pm0.05$ & $1.513\pm0.034$ \\ \cline{2-11}
 & Biomech & $0.996\pm0.000$ & $5.30\pm0.02$ & \bm{$1.132\pm0.029$} & $0.969\pm0.001$ & $2.77\pm0.02$ & $7.783\pm0.059$ & $0.992\pm0.001$ & $2.60\pm0.13$ & $1.378\pm0.035$ \\ \cline{2-11}
 & DeformIrisNet & $0.926\pm0.001$ & $2.15\pm0.01$ & $13.949\pm0.083$ & $0.814\pm0.001$ & $1.28\pm0.01$ & $26.409\pm0.099$ & $0.910\pm0.001$ &   $1.32\pm0.02$ & $16.037\pm0.091$ \\ \cline{2-11}
 & EyePreserve & \bm{$0.997\pm0.000$} & \bm{$5.72\pm0.02$}  & $1.463\pm0.035$ & $\bm{0.989\pm0.000}$ & $\bm{3.45\pm0.01}$ & \bm{$4.929\pm0.043$} & $\bm{0.993\pm0.000}$ &  $\bm{3.43\pm0.06}$ & \bm{$1.303\pm0.031$} \\ \midrule
\multirow{4}{*}{CIL} & Linear & $0.998\pm0.000$ & $4.87\pm0.01$ & $0.819\pm0.018$ & $0.984\pm0.001$ & $3.17\pm0.02$ & $6.492\pm0.044$ & $0.996\pm0.000$ & $2.60\pm0.03$ & $1.189\pm0.021$ \\ \cline{2-11}
 & Biomech & $0.998\pm0.000$ & $4.69\pm0.01$ & \bm{$0.706\pm0.017$} & $0.981\pm0.001$ & $3.04\pm0.02$ & $5.885\pm0.042$ & $\bm{0.997\pm0.000}$ & $\bm{2.91\pm0.04}$ & \bm{$0.867\pm0.015$} \\ \cline{2-11}
 & DeformIrisNet & $0.835\pm0.001$ & $1.43\pm0.00$ & $34.849\pm0.088$ & $0.818\pm0.001$ & $1.28\pm0.00$ & $26.842\pm0.082$ & $0.833\pm0.001$ & $1.19\pm0.01$ & $34.289\pm0.088$ \\ \cline{2-11}
 & EyePreserve & \bm{$0.999\pm0.000$} & \bm{$4.96\pm0.01$} & $0.995\pm0.019$ & $\bm{0.993\pm0.002}$ & $\bm{3.69\pm0.01}$ & \bm{$3.615\pm0.034$} & $\bm{0.997\pm0.000}$ & $2.50\pm0.04$ & $1.213\pm0.020$ \\ \midrule
\multirow{4}{*}{WBSPI} & Linear & $0.939\pm0.001$ & $2.30\pm0.02$ & $12.447\pm0.218$ & $0.834\pm0.002$ & $1.45\pm0.01$ & $23.999\pm0.265$ & $0.969\pm0.001$ & $1.12\pm0.02$ & $5.361\pm0.165$ \\ \cline{2-11}
 & Biomech & $0.940\pm0.001$ & $2.26\pm0.01$ & $12.573\pm0.202$ & $0.854\pm0.002$ & $1.50\pm0.01$ & \bm{$22.302\pm0.233$} & \bm{$0.970\pm0.001$} & $1.12\pm0.02$ & \bm{$5.103\pm0.141$} \\ \cline{2-11}
 & DeformIrisNet & $0.858\pm0.002$ & $1.52\pm0.01$ & $22.383\pm0.231$ & $0.740\pm0.002$ & $0.94\pm0.01$ & $31.925\pm0.215$ & $0.870\pm0.002$ & $0.75\pm0.01$ & $20.275\pm0.206$ \\ \cline{2-11}
 & EyePreserve & \bm{$0.950\pm0.001$} & \bm{$2.33\pm0.02$} & \bm{$11.309\pm0.200$} & \bm{$0.859\pm0.002$} & \bm{$1.62\pm0.01$} & $22.659\pm0.238$ & $0.966\pm0.001$ & \bm{$2.19\pm0.03$} & $5.378\pm0.149$ \\ \bottomrule
\end{tabular}%
}
\end{table*}

\begin{figure*}[!htbp] 
    \centering
    \begin{subfigure}{\textwidth}
        \centering
        \includegraphics[width=\linewidth]{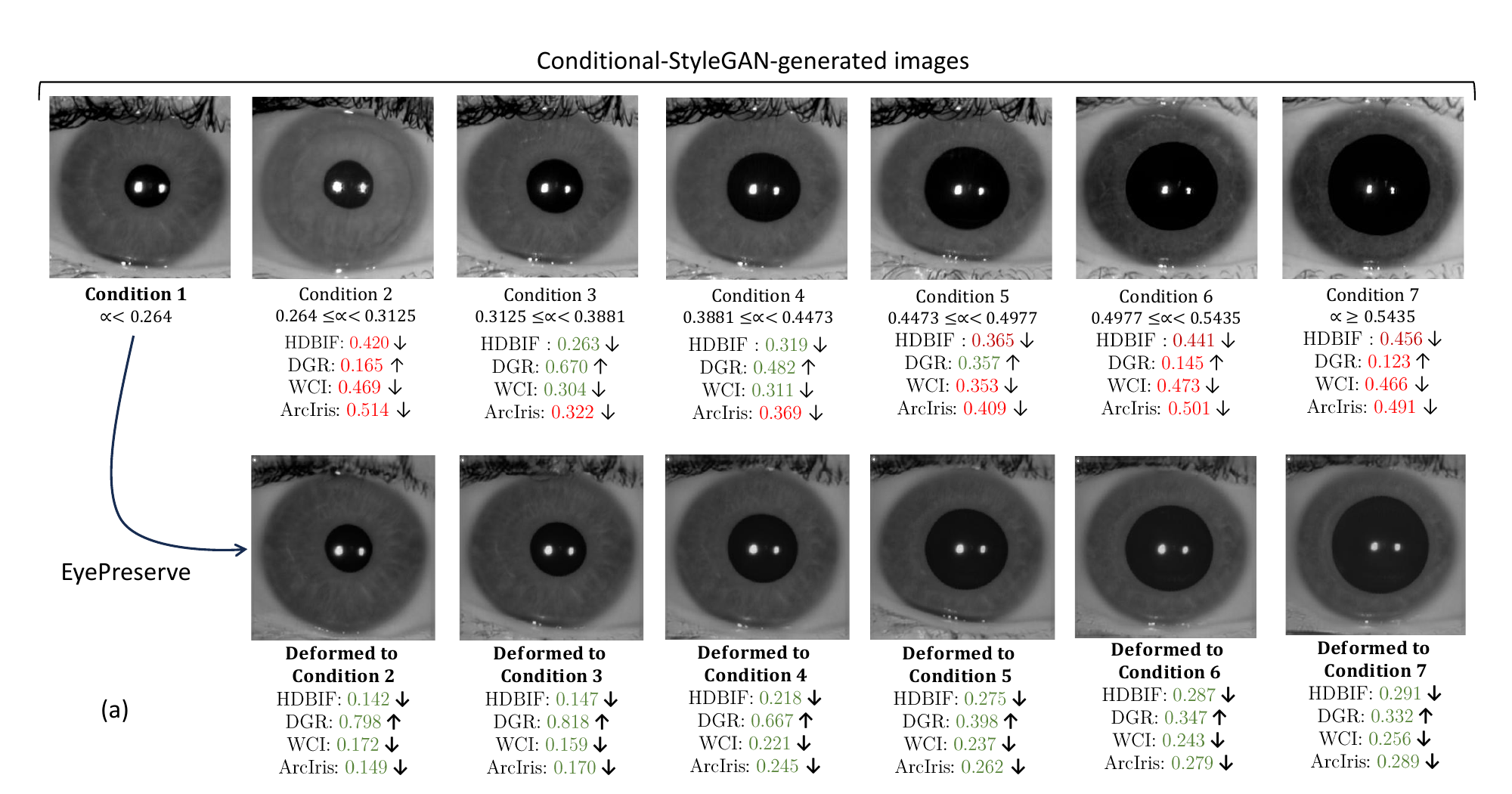}
        \captionlistentry{}
        \label{fig:conditional_comparison}
        \vspace{-10pt}
    \end{subfigure}
    \begin{subfigure}{\textwidth}
        \centering
        \includegraphics[width=\linewidth]{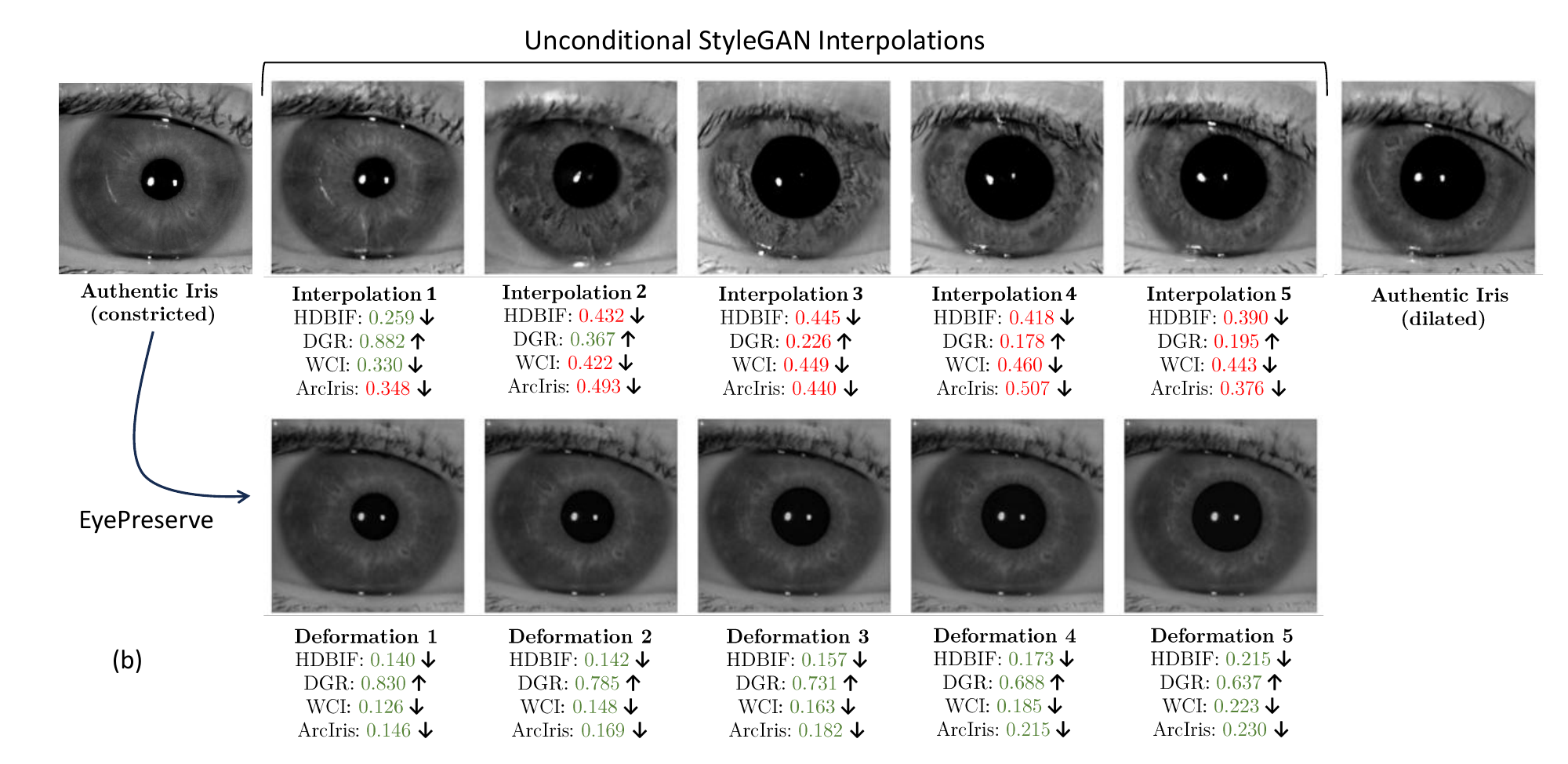}
        \captionlistentry{}
        \label{fig:unconditional_comparison}
    \end{subfigure}
    \vspace{-10pt}
    \setlength{\belowcaptionskip}{-7pt}
    \caption{\siamul{Illustration of identity preservation challenges in state-of-the-art StyleGAN models. HDBIF-based distance, WCI-based distance, and DGR-based similarity scores are shown for each comparison ($\uparrow$ denotes higher is better for similarity metrics; $\downarrow$ denotes lower is better for distance metrics). Green text indicates a match, and red text indicates a non-match. (a) A conditional StyleGAN fails to preserve identity when the same latent vector is used with different pupil-size conditions. We intentionally select a sample here where identity is preserved in a subset of conditions (conditions 3 and 4). (b) An unconditional StyleGAN also fails, as interpolating in the latent space between a constricted and dilated version of the same iris produces intermediate images that do not match. In both scenarios, the proposed {\it EyePreserve} model successfully deforms the original iris while preserving identity, as evidenced by the high match scores.}}
    \label{fig:GAN_vs_EyePreserve}
\end{figure*}

\subsection{Identity-Preservation in Iris Synthesis: StyleGAN vs. EyePreserve}
\label{sec:GAN_do_not_work}

To compare the {\it EyePreserve} model with state-of-the-art generative approaches, this subsection presents experiments with conditional and unconditional StyleGAN-based approaches, which -- while able to start from the GAN-inverted sample and synthesize iris images with varying pupil sizes -- are not able to preserve identity out of the box.

\vspace{4pt}\textit{Unconditional StyleGAN:} \siamulnew{For this experiment, we train a StyleGAN3 model on the CSOSIPAD and WBPD data. We utilize StyleGAN3 here because its translation equivariance prevents ``texture sticking'' during continuous spatial interpolation. Despite this architectural advantage,} linearly interpolating between small- and large-pupil images in StyleGAN3's latent space $W$ fails to preserve identity as shown in Fig.~\ref{fig:unconditional_comparison}. While the endpoints represent the same iris, intermediate frames yield fake, non-matching textures. In stark contrast, the proposed {\it EyePreserve} approach (bottom row in Fig.~\ref{fig:unconditional_comparison}) consistently maintains identity across varying pupil sizes, as reflected by the high comparison scores.


\vspace{4pt}\textit{Conditional StyleGAN:} \siamulnew{For this experiment, we train a conditional StyleGAN2-ADA model~\cite{karras2020training} on the class-balanced WBPD data. Unlike the unconditional interpolation experiment, this discrete generation task prioritizes overall image sharpness over continuous spatial transformations, making the crisper outputs of StyleGAN2 preferable to StyleGAN3.} To control the pupil size of the generated images, we introduced class conditions based on the pupil-to-iris ratio. Specifically, images with a pupil-to-iris ratio ranging from 0.2 to 0.7 were divided into seven discrete bins of approximately equal size. These bins were determined by sorting the dataset by $\alpha$ and distributing images equally into the following ranges: $[0.188, 0.264]$, $(0.264, 0.3125]$, $(0.3125, 0.3881]$, $(0.3881, 0.4473]$, $(0.4473, 0.4977]$, $(0.4977, 0.5435]$, and $(0.5435, \infty)$.  Each bin was assigned a unique class label. Our results indicate that conditional GANs frequently struggle to preserve identity. Furthermore, these models tend to generate synthetic iris images that lean toward identities encountered during training \cite{tinsley2022haven}. As illustrated in Fig. \ref{fig:conditional_comparison} upper row, the conditional StyleGAN failed to preserve identity when the same latent code was projected across these different $\alpha$ conditions. In contrast, the {\it EyePreserve} model (bottom row) demonstrated superior identity preservation, yielding significantly higher matching scores across the full spectrum of pupil sizes. 

To quantify identity preservation results, we compared the conditional StyleGAN2-ADA against {\it EyePreserve} using 1,000 unique latent codes across the seven distinct $\alpha$ conditions (as defined in Figure~\ref{fig:conditional_comparison}). We generated separate images for each class directly from the latent codes. In contrast, for the {\it EyePreserve} evaluation, we selected a single reference image from the StyleGAN-generated images in the median class (Condition 4: $0.3881 \leq \alpha < 0.4473$) for each latent code. We applied {\it EyePreserve} to synthesize the remaining six conditions. As shown in Table~\ref{tab:gan_v_ep}, the conditional StyleGAN fails to preserve identity, evidenced by low AUC and $d'$ scores. The {\it EyePreserve} model, however, robustly preserves identity across varying $\alpha$ values.

\begin{table}[!htbp]
\centering
\caption{\label{tab:gan_v_ep}\siamul{Identity preservation comparison: Measured by AUC, $d'$, and EER across varying pupil-to-iris ratios ($\alpha$), EyePreserve significantly outperforms conditional StyleGAN2-ADA on all three iris matchers (ArcIris, DGR, WCI).}}
\begin{tabular}{cccc}
\\ \toprule
    Matchers  & Metric  & Conditional StyleGAN & EyePreserve \\  \midrule
    \multirow{3}{*}{ArcIris}         &      AUC       &       $0.737\pm0.002$       &   $0.994\pm0.000$       \\ \cline{2-4}
                                 &      $d'$       &       $0.938\pm0.007$      &  $4.570\pm0.027$       \\ \cline{2-4}
                                 &      EER        &       $33.727\pm0.186$       &  $1.639\pm0.037$      \\ \cline{1-4}
    \multirow{3}{*}{DGR} &      AUC       &       $0.785\pm0.002$       &      $0.983\pm0.001$    \\ \cline{2-4}
                                 &      $d'$       &       $1.030\pm0.006$       &      $3.178\pm0.017$    \\ \cline{2-4}
                                 &      EER        &       $29.390\pm0.180$      &      $4.800\pm0.090$      \\ \cline{1-4}
    \multirow{3}{*}{WCI} &      AUC       &       $0.770\pm0.002$       &      $0.995\pm0.001$    \\ \cline{2-4}
                         &      $d'$       &       $0.860\pm0.013$       &      $3.831\pm0.071$\\  \cline{2-4} 
                         &      EER        &       $31.330\pm0.185$     &      $1.646\pm0.029$\\ \bottomrule        
\end{tabular}
\end{table}

\begin{figure*}[!htbp]
\centering
\includegraphics[width=\linewidth]{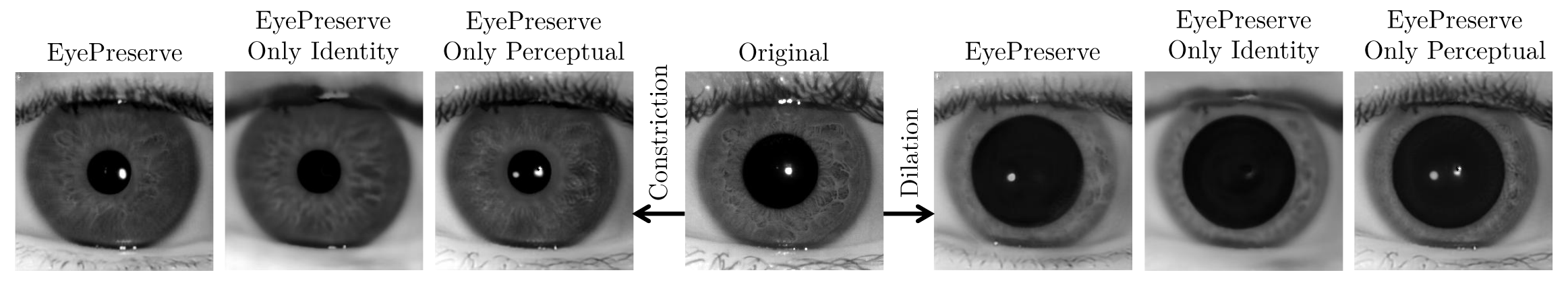}
\vspace{-10pt}
\setlength{\belowcaptionskip}{-7pt}
\caption{Visualization of synthesized iris images with different $\Delta$ (difrerence in pupil-to-iris ratios) for {\it EyePreserve} trained in three different scenarios: 1) with all of the losses, 2) with only the identity losses ($\mathcal{L}_{ID-F}$, $\mathcal{L}_{AE}$ and $\mathcal{L}_{D}$), and 3) with only the perceptual losses ($\mathcal{L}_{LPIPS}$, $\mathcal{L}_{MS-SSIM}$, $\mathcal{L}_{ISO-sharpness}$ and $\mathcal{L}_{patch-based}$). Using only the identity losses causes the model to focus mostly on lower frequencies, resulting in blurrier images. The opposite happens when only perceptual losses are used: the model focuses on higher frequencies, which leads to sharp images but can sometimes miss low-frequency identity-related features.}
\label{fig:ablation}
\end{figure*}

\subsection{\siamulnew{Conformance to the ISO/IEC 29794-6 Standard}}
\siamulnew{
We evaluated the biometric quality of the {\it EyePreserve}-deformed images by measuring their compliance with the ISO/IEC 29794-6 iris image quality standard~\cite{isostandardiris}. Because the {\it EyePreserve} model natively generates $256\time256$ images, we padded each sample with a clipped boundary to achieve a 4:3 aspect ratio, then resized them to $640\times480$. As detailed in Table~\ref{tab:iso_metrics}, the {\it EyePreserve} images perform comparably to the original dataset samples across a comprehensive suite of ISO quality metrics calculated using the MITRE BIQTIris tool~\cite{biqtiris}. While performance across most metrics appears to be the same, we do observe a reduction in the sharpness score for the generated images. This decrease occurs because the model's training focuses primarily on the iris region. Aside from MSE (a loss inherently prone to smoothing high-frequency textures), all other losses are calculated on the masked polar-normalized iris images. Consequently, details in peripheral features such as eyelashes become blurred, which inherently lowers the global ISO sharpness metric evaluated across the entire image. However, the highly similar scores observed across all other metrics, notably the average Overall Quality score of $97.15 \pm 14.43$ observed in the {\it EyePreserve}-generated iris images closely matching the overall quality score of $96.63 \pm 17.67$ observed in the real iris images, suggest that {\it EyePreserve}-sourced samples are ISO/IEC 29794-6-compliant, offering the high fidelity required for reliable integration into iris biometric systems. We include the box plots for all these metrics in our supplementary materials.

\begin{table}[htpb]
\centering
\caption{Comparison of ISO/IEC 29794-6 Metrics (Mean $\pm$ Standard Deviation)}
\label{tab:iso_metrics}
\resizebox{\columnwidth}{!}{
\begin{tabular}{lcc}
\toprule
 Metric & Original & EyePreserve \\
\midrule
Grayscale Utilization & $7.22 \pm 0.25$ & $7.27 \pm 0.24$ \\
Iris Pupil Concentricity & $72.94 \pm 12.43$ & $76.17 \pm 11.79$ \\
Iris Pupil Contrast & $82.95 \pm 7.02$ & $81.45 \pm 8.66$ \\
Iris Pupil Ratio & $30.91 \pm 16.24$ & $28.35 \pm 15.06$ \\
Iris Sclera Contrast & $86.87 \pm 31.62$ & $84.15 \pm 33.70$ \\
Margin Adequacy & $97.99 \pm 7.75$ & $98.45 \pm 6.73$ \\
Pupil Boundary Circularity & $88.05 \pm 12.73$ & $84.65 \pm 17.14$ \\
Sharpness & $59.94 \pm 28.26$ & $15.28 \pm 11.69$ \\
Usable Iris Area & $87.27 \pm 16.45$ & $90.36 \pm 14.38$ \\
Normalized Grayscale Utilization & $1.00 \pm 0.00$ & $1.00 \pm 0.00$ \\
Normalized Iris Diameter & $1.00 \pm 0.00$ & $1.00 \pm 0.00$ \\
Normalized Iris Pupil Concentricity & $1.00 \pm 0.00$ & $1.00 \pm 0.00$ \\
Normalized Iris Pupil Contrast & $1.00 \pm 0.00$ & $1.00 \pm 0.00$ \\
Normalized Iris Pupil Ratio & $1.00 \pm 0.00$ & $1.00 \pm 0.00$ \\
Normalized Iris Sclera Contrast & $1.00 \pm 0.03$ & $1.00 \pm 0.04$ \\
Normalized Margin Adequacy & $0.97 \pm 0.17$ & $0.98 \pm 0.13$ \\
Normalized Sharpness & $1.00 \pm 0.00$ & $0.99 \pm 0.05$ \\
Normalized Usable Iris Area & $0.96 \pm 0.14$ & $0.98 \pm 0.12$ \\  \midrule 
Overall Quality & $96.63 \pm 17.67$ & $97.15 \pm 14.43$ \\
\bottomrule
\end{tabular}
}
\end{table}
\vspace{-12pt}
}

\subsection{Assessment of Human Perceptual Realism}
To quantitatively assess the visual realism of the iris images generated by our synthesis pipeline, we computed the Fréchet Inception Distance (FID)~\cite{heusel2017gans}, Kernel Inception Distance (KID)~\cite{binkowski2018demystifying}, and CLIP Maximum Mean Discrepancy (CMMD)~\cite{jayasumana2024rethinking} between 1,479 real iris images randomly sampled from WBPD and HWS test sets, and 1,479 {\it EyePreserve}-generated images. We observed FID of 69.10, KID of 0.0622, and a CMMD of 0.1366.

While FID is the most commonly used metric to measure perceptual realism, it is a biased estimator that often over-penalizes models when evaluated on smaller datasets or specialized domains such as iris biometrics. Unlike FID, KID is an unbiased estimator as it measures the squared Maximum Mean Discrepancy (MMD) between the Inception-v3 feature distributions. This means its expected value does not depend on the sample size. In addition, KID does not assume that the image features follow a multivariate normal distribution, making it a more reliable metric for the complex, non-Gaussian textures found in iris patterns. Our achieved score of 0.062 indicates a stable and accurate alignment between the generated and real data distributions.

\siamulnew{However, same as FID, KID still relies on Inception-v3 features trained on ImageNet, which may not adequately represent the unique structural characteristics of human irises. Therefore, to address this limitation, we also utilize the state-of-the-art CLIP Maximum Mean Discrepancy (CMMD) measure, which instead utilizes CLIP embeddings.} These embeddings are trained on a significantly more diverse dataset of image-text pairs, allowing for a feature space that is more closely aligned with human perception. A CMMD score of 0.1366 confirms that {\it EyePreserve}-synthesized images maintain high semantic fidelity and structural consistency with authentic iris samples.

For completeness, we calculated the visual realism metrics between 1,479 randomly sampled images from the StyleGAN2-ADA model and the same 1479 real iris images as well. The StyleGAN model achieved an FID of 91.95, a KID of 0.0813, and a CMMD of 0.2436, which are consistently worse than those for the {\it EyePreserve} model. 


\subsection{Ablation Study}
\label{sec:ablation_study}
To better understand how the components of identity preservation loss and the components of perceptual loss interact and affect the image generation process, we carry out an ablation study, in which we train our model with only the identity losses ($\mathcal{L}_{ID-F}$, $\mathcal{L}_{AE}$ and $\mathcal{L}_{D}$) and only the perceptual losses ($\mathcal{L}_{LPIPS}$, $\mathcal{L}_{MS-SSIM}$, $\mathcal{L}_{ISO-sharpness}$ and $\mathcal{L}_{patch-based}$).

\begin{table}[!htbp]
\centering
\caption{\label{tab:ablation}\siamul{Comparison of AUCs and EERs for ArcIris, DGR and WCI matchers obtained for the {\it EyePreserve} model trained in three different scenarios: 1) with all of the losses, 2) with only the identity losses ($\mathcal{L}_{ID-F}$, $\mathcal{L}_{AE}$ and $\mathcal{L}_{D}$), and 3) with only the perceptual losses ($\mathcal{L}_{LPIPS}$, $\mathcal{L}_{MS-SSIM}$, $\mathcal{L}_{ISO-sharpness}$ and $\mathcal{L}_{patch-based}$).}}
\resizebox{\columnwidth}{!}{%
\begin{tabular}{cllcccc}
\toprule
\multicolumn{2}{c}{\multirow{2}{*}{\bf Dataset}} & \multirow{2}{*}{\bf Matcher} & \multirow{2}{*}{\bf Metric} & \bf All & \bf Only & \bf Only \\
 & & & & \bf Losses & \bf Identity & \bf Perceptual \\ \midrule
\multirow{12}{*}{HWS} & \multirow{6}{*}{$\Delta \leq 0.1$} & \multirow{2}{*}{ArcIris} & AUC & $0.9964$   &    $0.9964$    &   $0.9946$      \\ \cline{4-7}
 & & & EER($\%$) & $0.6370$ & $0.9250$ & $0.9413$ \\ \cline{3-7} 
 & & \multirow{2}{*}{DGR} & AUC & $0.9913$   & $0.9816$      & $0.9746$        \\ \cline{4-7}
 & & & EER($\%$) & $1.6863$ & $1.3133$ & $1.8295$ \\ \cline{3-7} 
 & & \multirow{2}{*}{WCI} & AUC &  $0.9914$  &  $0.9909$     &  $0.9901$ \\ \cline{4-7}
 & & & EER($\%$) & $1.2857$ & $1.8766$ & $2.8901$ \\ \cline{2-7}
 & \multirow{6}{*}{$\Delta \geq 0.2$} & \multirow{2}{*}{ArcIris} & AUC & $0.9991$   &   $0.9986$    &   $0.9967$      \\ \cline{4-7}
 & & & EER($\%$) & $0.6788$ & $1.1340$ & $1.2924$ \\ \cline{3-7} 
 & & \multirow{2}{*}{DGR} & AUC & $0.9890$   & $0.9779$      & $0.9396$        \\ \cline{4-7}
 & & & EER($\%$) & $4.1073$ & $2.8470$ & $5.4896$\\ \cline{3-7}
 & & \multirow{2}{*}{WCI} & AUC & $0.9824$  &   $0.9798$    &    $0.9724$    \\ \cline{4-7}
 & & & EER($\%$) & $2.9750$ & $3.1540$ & $4.3107$ \\ \midrule
\multicolumn{2}{c}{\multirow{6}{*}{Q-FIRE}} & \multirow{2}{*}{ArcIris} & AUC & $0.9968$   & $0.9949$      &   $0.9940$      \\ \cline{4-7}
 & & & EER($\%$) & $1.4635$ & $1.9879$ & $2.1401$ \\ \cline{3-7} 
 & & \multirow{2}{*}{DGR} & AUC & $0.9891$   & $0.9652$      & $0.9497$        \\ \cline{4-7}
 & & & EER($\%$) & $4.9286$ & $11.3499$ & $13.4339$ \\ \cline{3-7}
 & & \multirow{2}{*}{WCI} & AUC & $0.9927$ & $0.9917$ & $0.9912$ \\ \cline{4-7}
 & & & EER($\%$) & $1.3048$ & $2.2692$ & $2.5093$\\ \midrule
\multicolumn{2}{c}{\multirow{6}{*}{CIL}} & \multirow{2}{*}{ArcIris} & AUC & $0.9990$   &  $0.9975$     &  $0.9972$       \\ \cline{4-7}
 & & & EER($\%$) & $0.9932$ & $1.5049$ & $1.7651$ \\ \cline{3-7} 
 & & \multirow{2}{*}{DGR} & AUC & $0.9931$   & $0.9881$      & $0.9780$        \\ \cline{4-7}
 & & & EER($\%$) & $3.6148$ & $1.9481$ & $2.9362$\\ \cline{3-7}
 & & \multirow{2}{*}{WCI} & AUC & $0.9971$ & $0.9948$ & $0.9936$ \\ \cline{4-7}
 & & & EER($\%$) & $1.2133$ & $2.8319$ & $2.1675$\\ \bottomrule
\end{tabular}%
}
\end{table}

Figure~\ref{fig:ablation} illustrates how different sets of losses affect the iris image synthesis process, and Table~\ref{tab:ablation} presents the results of iris recognition performance evaluation for each composition of the loss function. 
Using all losses and only identity losses results in about the same iris recognition performance. However, generated images are blurrier since iris biometric features are usually located in lower frequency ranges. Also, we usually see a slight drop in performance when using only perceptual losses, since these losses focus on spatial image frequencies that are important from a human perception perspective, which may be different from those used by iris recognition algorithms. We thus conclude that to obtain synthetic iris images that are usable in both automatic and human-based iris recognition, it is beneficial to use the entire set of loss components, as illustrated in Fig.~\ref{fig:train}.

\subsection{Inference Time}

Without batch processing, the {\it EyePreserve} model takes ~50.24 ms per image on a single NVIDIA RTX 3080 Mobile GPU (AMD Ryzen 7 5800H CPU). Batching would significantly accelerate this process.

\section{Selected Applications Beyond Iris Matching}

\subsection{Rectification of Diseased-Deformed Irises}

\begin{figure*}[!htbp]
\centering
\includegraphics[width=\linewidth]{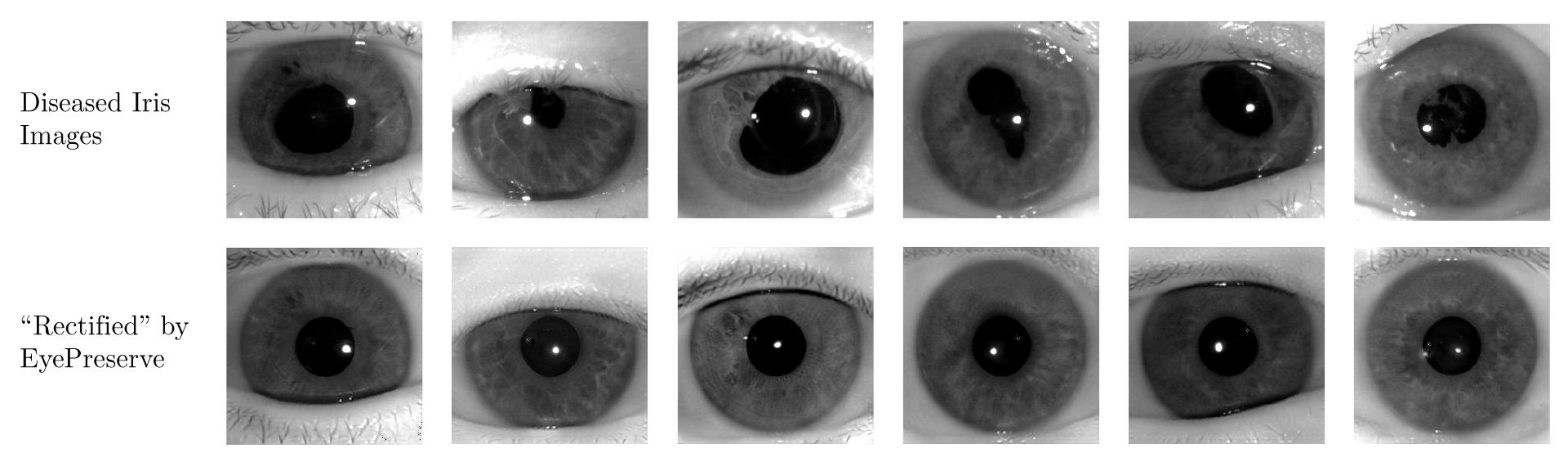} 
\setlength{\belowcaptionskip}{-7pt}
\caption{The {\it EyePreserve} model can rectify geometrical deformations of the iris texture caused by diseases, and make the pupil shape circular while preserving the subject's identity features. Samples of diseased eye images have been sourced from \cite{trokielewicz2015assessment}.}
\label{fig:diseasefix}
\end{figure*}

Interestingly, the {\it EyePreserve} model, although never trained for that purpose, can deform any shape of an iris to any other arbitrary shape, maximizing the preservation of identity-related features. That opens new, previously non-existent, applications for the iris-specific image-to-image translation model as {\it EyePreserve}. One of them is ``rectification'' of iris images, which -- due to diseases -- have irregular pupil shapes. Fig.~\ref{fig:diseasefix} visualizes a few original samples from the {\it Warsaw-BioBase-Disease-Iris v2.1} dataset \cite{trokielewicz2015assessment} representing eyes that suffer from geometric deformations of the pupil, and rectified images by {\it EyePreserve}. The target masks for the rectified images were estimated by (a) finding the outer iris boundary in the original (diseased iris) image, (b) cutting a pupil circle concentric with the iris circle and ensuring a population-average pupil-to-iris ratio $\alpha=0.35$, and (c) using the ``inside-eyelid'' mask (cf. Fig. \ref{fig:mask_estimation}) to remove the remaining non-iris regions.

\begin{figure}[!ht]
\centering
\includegraphics[width=\linewidth]{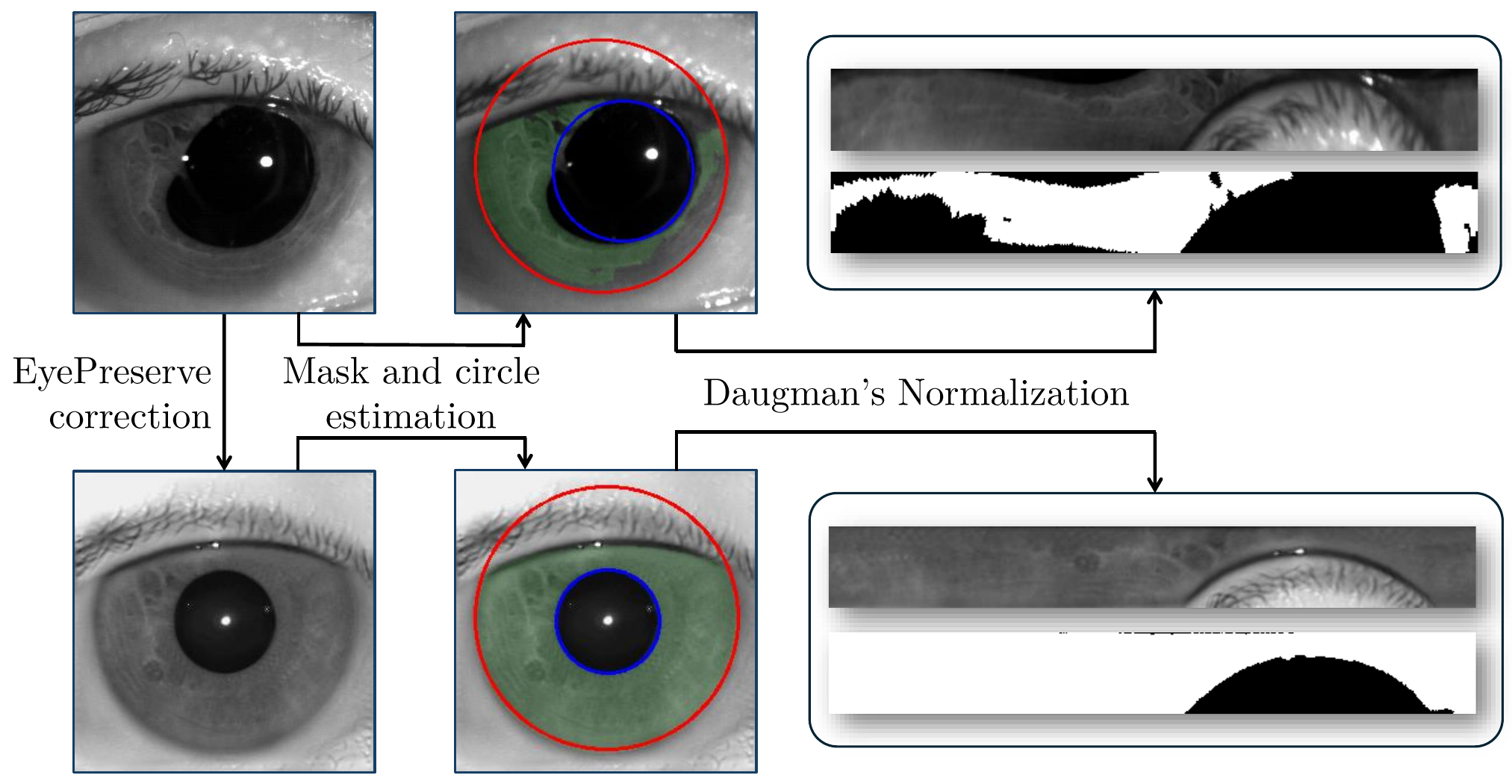}
\setlength{\belowcaptionskip}{-7pt}
\caption{\siamul{Normalization workflow for diseased vs. {\it EyePreserve}-corrected images. The first row shows the distortion in the polar transformation caused by an irregular pupil shape in the diseased eye. The second row demonstrates how {\it EyePreserve} ensures circularity of iris boundaries, resulting in regular polar iris images, enabling more accurate biometric comparisons by matchers following Daugman's approach.
}}
\label{fig:disease_vs_fix_polarnorm}
\end{figure}

Such rectification can be useful in human examination of iris images, especially for pairs of images representing irises before and after surgeries altering the iris structure or pupil shape. To assess whether this rectification preserves identity, we performed an evaluation using ArcIris, DGR, and WCI matchers, and compared the performance obtained for the original (not rectified) images and for those after rectification. For all of our matchers, we require normalized images as input. \siamul{Figure~\ref{fig:disease_vs_fix_polarnorm} illustrates how this normalization looks for original diseased and {\it EyePreserve}-rectified images. We experiment with a range of $\alpha$ values between 0.2 and 0.7 with a step size of 0.05, and find that an $\alpha$ of 0.3 provides the best results.} As shown in Table~\ref{tab:correction}, such rectification can be performed while preserving identity. Moreover this method
achieved better performance for $\alpha \leq 0.3$ with the WCI matcher.

We do not see similar improvements in the case of DGR, which creates a graph in which the iris features are graph nodes and the spatial relations between these features are edges. The irregular pupil shape gives rise to ``strong'' but fake features for the model, which uses them as nodes for its graph. When we rectify these irregularities, these ``strong'' characteristics are lost, leading to a slight and misleading performance loss. \siamul{We see a similar trend in ArcIris. However, the performance difference is even smaller. ArcIris is a neural network that runs on the polar-normalized iris image and is free to focus on any features. Thus, similar to DGR, irregularities in the original image can work as ``strong'' but fake features for this model.}

\begin{table}[!htbp]
\centering
\caption{\label{tab:correction} Comparison of AUCs obtained before and after {\it EyePreserve}-based rectifications of geometric deformations caused by diseases (examples of such rectifications are shown in Fig. \ref{fig:diseasefix}). The ``geometry'' subset of the dataset of images of diseased eyes~\cite{trokielewicz2015assessment} was used for this evaluation. 
\siamul{$\alpha$ is the requested pupil-to-iris ratio of the ``rectified'' iris image from {\it EyePreserve}. 
}}
\begin{tabular}{llcc}
\toprule
Method                 & Metric & \begin{tabular}[c]{@{}c@{}}Original\\ Diseased\\ ``Geometry"\\ Set\end{tabular} & \begin{tabular}[c]{@{}c@{}}Rectified by\\ EyePreserve\\ (target $\alpha=0.3$)\end{tabular} \\ \hline
\multirow{3}{*}{ArcIris} & AUC    &    $0.992\pm0.001$                                                                 &      $0.989\pm0.001$                                                                           \\ \cline{2-4}
                       & $d'$   &        $3.474\pm0.047$                                                                 &  $3.244\pm0.045$                                                                               \\ \cline{2-4}
                       & EER    &    $1.067\pm0.522$                                                                                &      $3.899\pm0.884$                                                                                         \\ \hline
\multirow{3}{*}{DGR}   & AUC    &              $0.980\pm0.002$                                                          &                       $0.966\pm0.002$                                                         \\ \cline{2-4}
                       & $d'$   &       $2.749\pm0.044$                                                               &                $2.795\pm0.040$                                                               \\ \cline{2-4}
                       & EER    &              $6.581\pm1.118$                                                         &             $8.727\pm1.154$                                                                             \\ \hline
\multirow{3}{*}{WCI}   & AUC    &    $0.952\pm0.001$                                                                     &   $0.989\pm0.002$                                                                              \\ \cline{2-4}
                       & $d'$   &   $0.681\pm0.010$                                                                      &   $2.742\pm0.037$                                                                              \\ \cline{2-4}
                       & EER    &                             $6.033\pm0.753$                                                       &             $3.786\pm0.867$                                                                                 \\ \bottomrule                                                                                               
\end{tabular}
\end{table}

\subsection{Morphing Iris Images into Irregular Pupil Shapes}

\begin{figure*}[!htbp]
\centering
\includegraphics[width=0.94\linewidth]{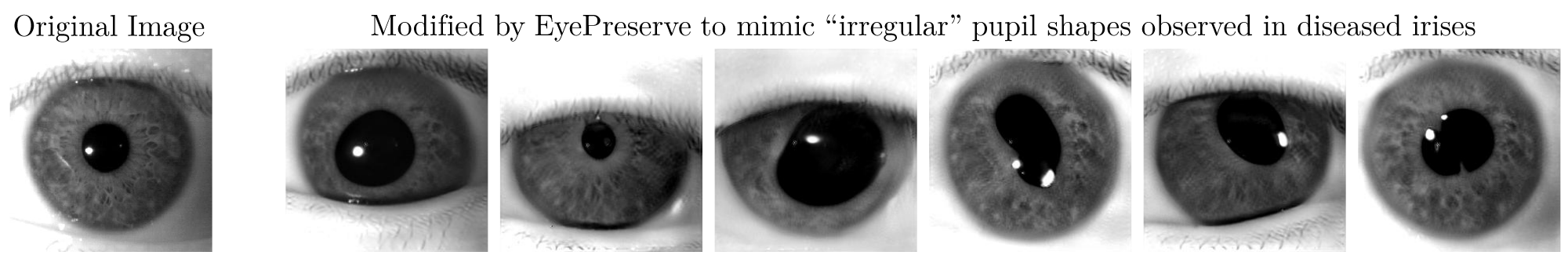}
\setlength{\belowcaptionskip}{-7pt}
\caption{Opposite to what was demonstrated in Fig. \ref{fig:diseasefix}, the {\it EyePreserve} model can modify healthy iris images to mimic irregular pupil shapes observed in diseased irises. The target image mask is the one estimated from an iris suffering from a disease causing pupil shape distortions (in the example, masks were estimated for samples shown in Fig.~\ref{fig:diseasefix}).}
\label{fig:normaltodisease}
\end{figure*}

\begin{figure*}[!htbp]
\centering
\includegraphics[width=0.94\linewidth]{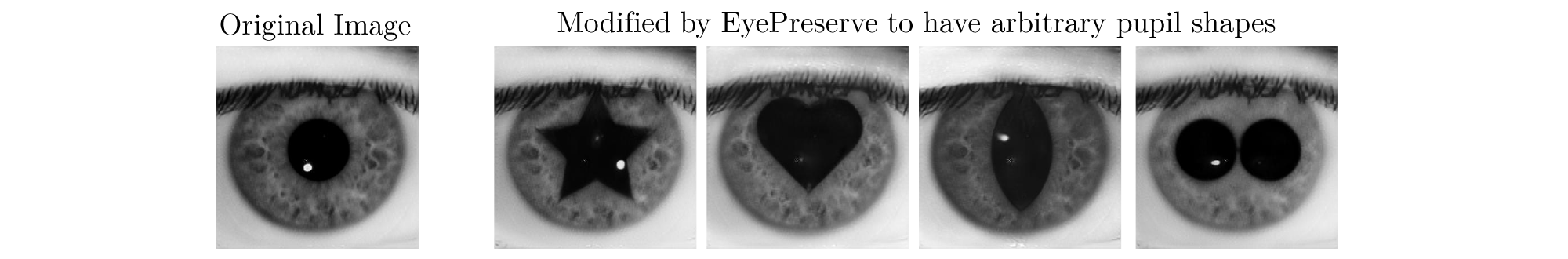}
\caption{Visualization of identity-preserving deformation of iris texture into arbitrary shapes.}
\label{fig:weirdanimations}
\end{figure*}

The {\it EyePreserve} model's capabilities can be extended to
deform the iris into arbitrary pupil shapes. Fig.~\ref{fig:normaltodisease} shows example results of modifying healthy iris images to mimic irregular pupil shapes that can be observed in diseased irises. Fig.~\ref{fig:weirdanimations} shows example deformations to arbitrary pupil shapes. \siamul{This feature of the {\it EyePreserve} model can be useful in synthesizing images of non-existent subjects mimicking selected eye diseases for training forensic iris examiners, and law enforcement labs collecting iris images for national identification systems, such as the FBI's Next Generation Identification (NGI) \cite{FBI_NGI_webpage}.}

%% file: sections/6_limitations.tex
\section{Limitations}
\label{sec:limitations}

Although {\it EyePreserve} advances non-linear, identity-preserving iris deformation, it has several practical limitations. The method relies on accurate segmentation of training images during data creation, and also when deforming iris images based solely on the pupil-to-iris ratio, because the target mask in this scenario is derived from the original mask, pupil, and iris circle parameters (see Fig.~\ref{fig:mask_estimation}).

\siamulnew{While {\it EyePreserve} demonstrates strong generative robustness to irregular segmentation masks by synthesizing plausible textures from extreme shapes (\eg, Figures~\ref{fig:diseasefix}, \ref{fig:normaltodisease}, and \ref{fig:weirdanimations}), downstream evaluation remains inherently tied to mask quality. In our bidirectional pair comparison methodology (shown in Figure~\ref{fig:dilate_constrict_comp}), each image is deformed to match the mask of its comparison counterpart before they are compared, and their scores are averaged. Because the masks actively define the target geometry for the cross-deformation, severe segmentation errors can propagate into the generative phase. This corrupts the synthesized iris texture and artificially degrades the comparison scores, regardless of the autoencoder's inherent generative robustness.}

Moreover, when the target mask exposes regions occluded in the input (\eg, larger eyelid opening), the model will ``hallucinate'' the iris texture in this region. This hallucinated texture is randomly generated (though there could be some extrapolation from the texture available at the boundaries), and there is no guarantee that it will match the original texture. This is, however, not specifically the limitation of {\it EyePreserve}, and rather a limitation of any iris image-to-image translation approach, in which missing texture needs to be guessed. One simple countermeasure is to annotate such hallucinated areas, \eg, by masking out these regions to exclude them from feature extraction.

There is also an inherent trade-off between identity-focused losses and perceptual/sharpness losses (identity-only training produces blurrier outputs, but these outputs are sometimes more biometrically stable, while perceptual losses give sharper but potentially less identity-faithful images). Similarly, when evaluating very small pupil differences, the autoencoder’s reconstruction cost can slightly reduce recognition scores compared to doing nothing, so applying the model blindly to all samples is not always optimal.

\siamulnew{Additionally, synthesizing non-existing identities relies on a foundational generative model (\eg, DDPM and StyleGAN3 in our pipeline), which introduces its own boundaries. The generator has a finite capacity to produce distinct identities (the unique combination of a subject ID and left/right eye side). Consequently, it cannot function as an unlimited data factory. Furthermore, prior studies have shown that StyleGAN architectures are susceptible to identity leakage~\cite{tinsley2022havent}, where generated images may inadvertently retain features from the training dataset. Since this generative module is independent of the core {\it EyePreserve} deformation model, future developments can swap in more robust generators to mitigate these capacity and leakage constraints.

Currently, {\it EyePreserve} is optimized specifically for NIR images as a single-channel model. There is a notable lack of large-scale, low-noise visible wavelength datasets focused specifically on isolated pupil dilation. This lack of data prevents the effective training of an RGB-native version of the {\it EyePreserve} model.}

Finally, {\it EyePreserve} was trained with healthy iris images with varying pupil sizes. It did not encounter highly irregular iris texture and pupil shapes (as observed in diseased eyes) during training, so while the rectification done using it can help matchers and human examiners, it does not guarantee that the reconstruction will represent the real disease-driven deformations.

%% file: sections/7_conclusions.tex
\section{Conclusions}
\label{sec:conclusions}

Complex deformations in the texture of the iris are difficult to model and require strong anatomical assumptions, which are often difficult to make. This paper proposes the first deep learning-based non-linear iris deformation model that generalizes well to unknown identities, along with a complete iris synthesis toolkit, which preserves identity when deforming the iris texture. The proposed model learns the visual appearance of the intricate iris muscle deformations without the need for often oversimplified anatomical assumptions and learns them quite effectively, offering better performance of tested iris recognition methods compared to state-of-the-art linear and non-linear (bio-mechanical-based) iris texture deformation models for significant differences in pupil dilation.

This work advances the iris recognition field in several ways: (a) offers better compensation of pupil dilation translating to better recognition performance compared to state-of-the-art models, especially in case of large differences in pupil size between the probe and gallery samples, (b) by correct alignment of iris features it serves better the emerging field of forensic human experts-based iris examination, (c) it offers a complete pipeline of identity-preserving iris image synthesis, potentially enhancing recognition and presentation attack detection datasets. 

\siamul{Future work will explore extending state-of-the-art diffusion models to iris synthesis, with a primary focus on identity preservation. Adapting these frameworks to handle iris deformation, specifically the complex task of varying pupil size, represents a non-trivial technical challenge. While GANs force identity into a small latent vector, diffusion-based models have no such bottleneck; the noise is as large as the image, meaning it doesn't ``compress'' identity, and rather increases the texture entropy. Thus, we require novel approaches to handle non-linear texture warping while maintaining biometric integrity within the diffusion training paradigm.}

%% file: biographies.tex
\begin{IEEEbiography}[{\includegraphics[width=1in, height=1.25in, clip, keepaspectratio]{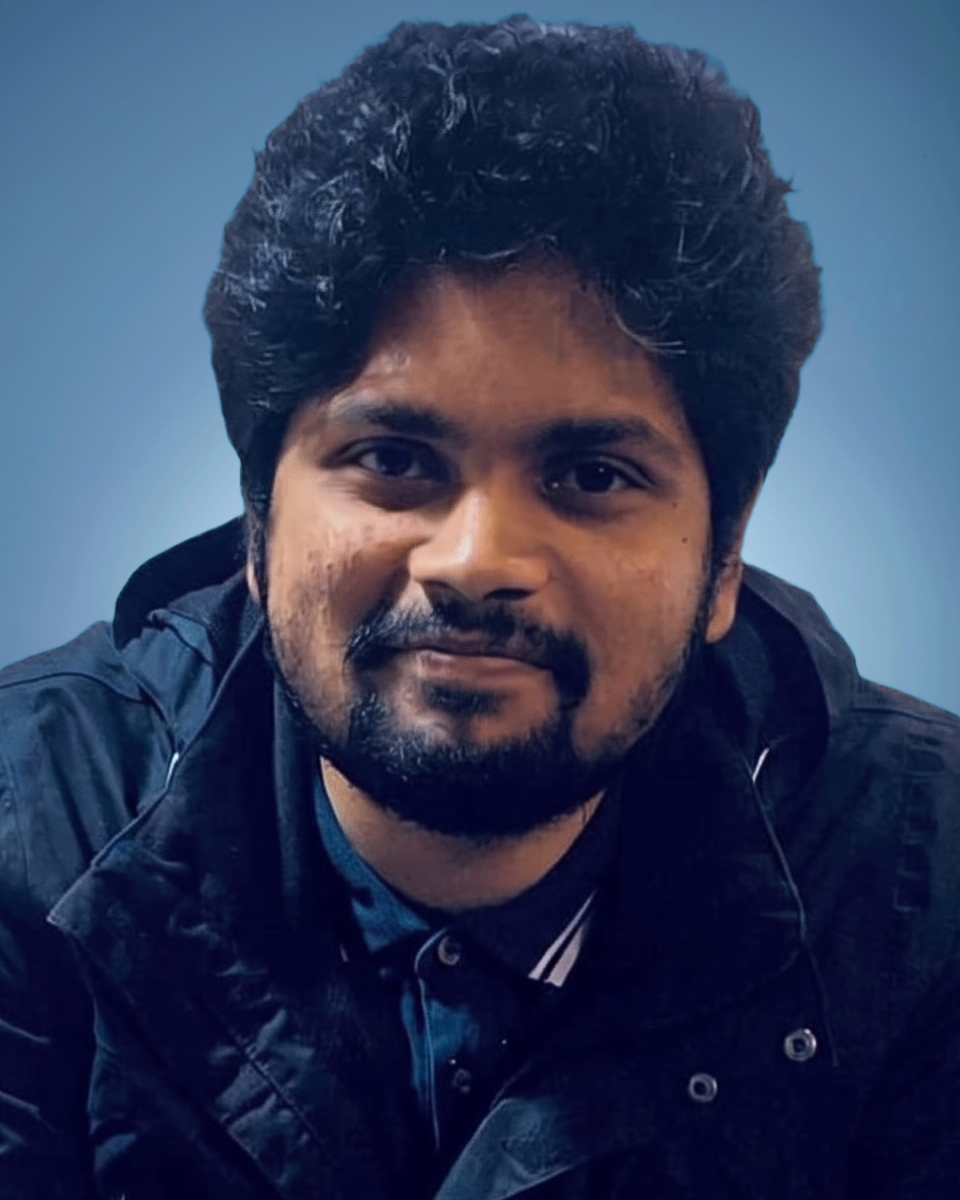}}]{Siamul Karim Khan}
received the B.S. degree in computer science and engineering from the Bangladesh University of Engineering and Technology (BUET), Dhaka, Bangladesh, in 2017. He is currently a Ph.D. candidate in computer science and engineering at the University of Notre Dame, Notre Dame, IN, USA. His research encompasses deep learning methods for iris segmentation and recognition, identity-preserving iris synthesis, and demographic fairness evaluations for iris recognition. Alongside his academic research, he gained professional experience through two internships at Meta, focusing on content ranking and multimodal evaluation. His research has been published in reputable venues including IEEE Transactions on Biometrics, Behavior, and Identity Science (T-BIOM), IEEE International Joint Conference on Biometrics (IJCB), IEEE/CVF Winter Conference on Applications of Computer Vision (WACV), Advances in Social Network Analysis and Mining (ASONAM), and Pacific-Asia Conference on Knowledge Discovery and Data Mining (PAKDD).
\end{IEEEbiography}

\begin{IEEEbiography}[{\includegraphics[width=1in, height=1.25in, clip, keepaspectratio]{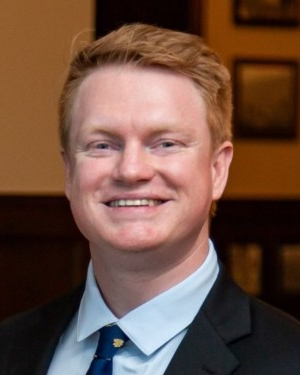}}]{Patrick Tinsley} 
received the Ph.D. degree in computer science from the University of Notre Dame, Notre Dame, IN, USA, where his doctoral dissertation focused on trust, artificial intelligence, and synthetic biometrics. As a researcher at the University of Notre Dame, his work focused on developing human-guided synthetic data detectors and privacy-safe generative models. He has authored and coauthored numerous articles in premier venues, including the IEEE Transactions on Artificial Intelligence (TAI), IEEE International Joint Conference on Biometrics (IJCB), IEEE/CVF Winter Conference on Applications of Computer Vision (WACV), and AAAI Conference on Artificial Intelligence (AAAI). He is currently the Head of AI research with AngelEye Health. His recent work focuses on revolutionizing neonatal intensive care (NICU) by integrating artificial intelligence and computer vision into bedside camera networks to enhance clinical decision-making, personalize patient care, and improve overall care coordination.
\end{IEEEbiography}

\begin{IEEEbiography}[{\includegraphics[width=1in, height=1.25in, clip, keepaspectratio]{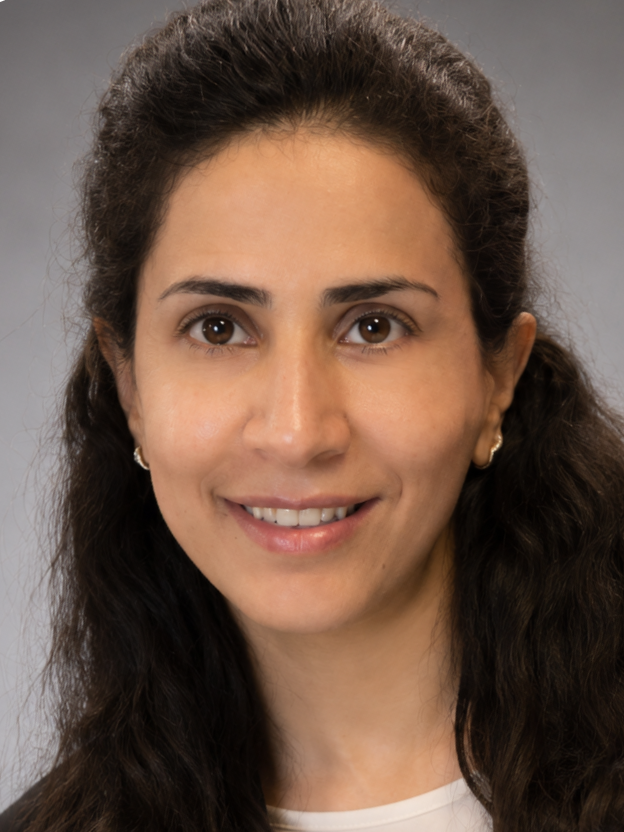}}]{Mahsa Mitcheff}
received the B.S. degree in hardware engineering from Islamic Azad University, Arak, Iran, in 2008. She is currently pursuing the Ph.D. degree in computer science and engineering at the University of Notre Dame, Notre Dame, IN, USA. Her research focuses on iris presentation attack detection. Her work investigates the use of synthetic iris images to improve presentation attack detection while preserving individual privacy. She has developed a controlled iris image augmentation framework by exploring the latent spaces of generative models, enabling privacy-safe data synthesis for robust and reliable biometric systems.
\end{IEEEbiography}

\begin{IEEEbiography}[{\includegraphics[width=1in, height=1.25in, clip, keepaspectratio]{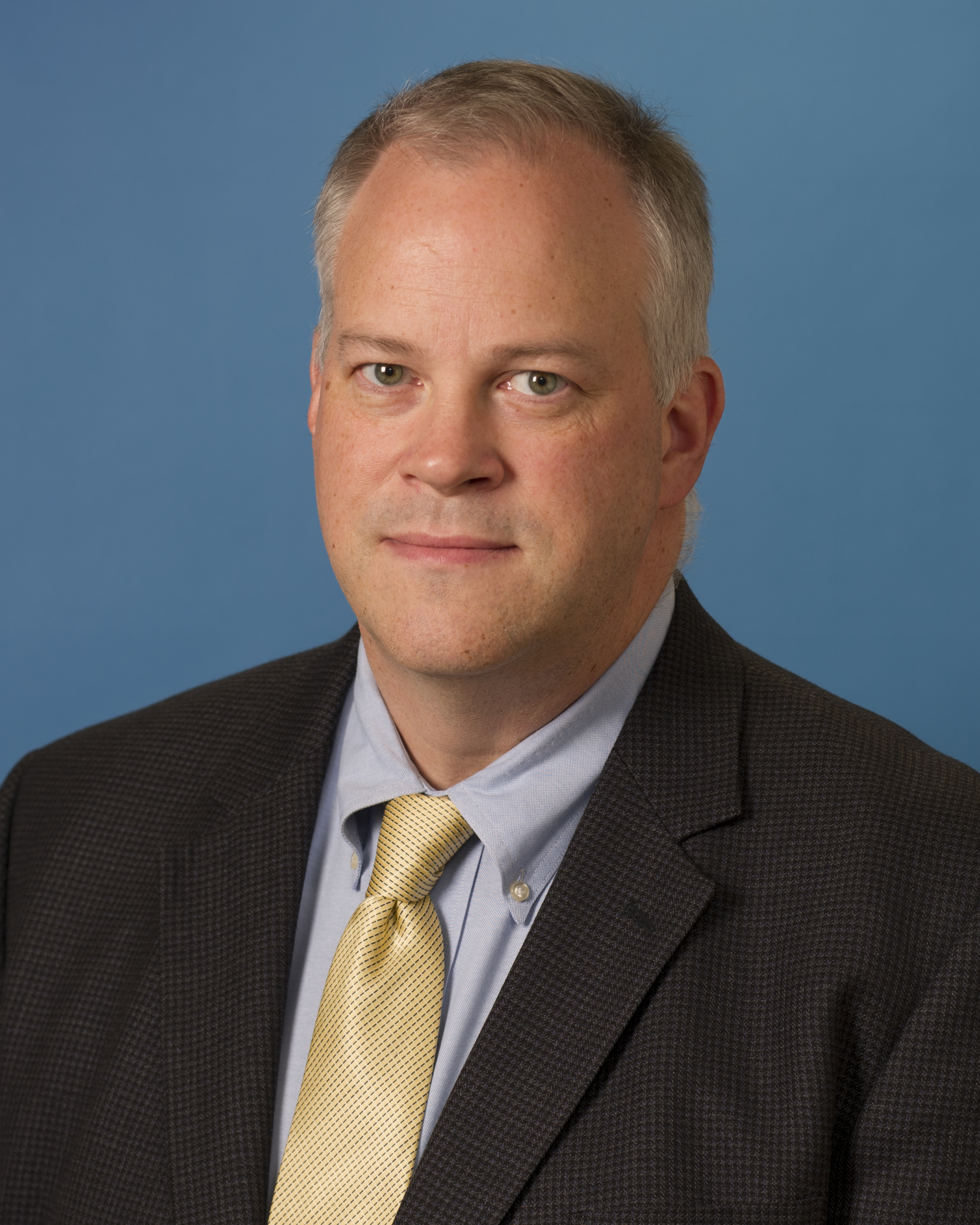}}]{Patrick J. Flynn}
is Professor Emeritus of Computer Science and Engineering at the University of Notre Dame.
He is an IEEE Fellow, an IAPR Fellow, and an ACM Distinguished Scientist. His research interests include computer vision and pattern recognition.
\end{IEEEbiography}

\begin{IEEEbiography}[{\includegraphics[width=1in, height=1.25in, clip, keepaspectratio]{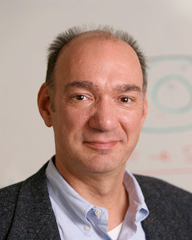}}]{Kevin W. Bowyer}
is the Schubmehl-Prein Family Professor Emeritus in Computer Science and Engineering at the University of Notre Dame.  Professor Bowyer is a Fellow of the American Academy for the Advancement of Science “for distinguished contributions to the field of computer vision and pattern recognition, biometrics, object recognition and data science”, a Fellow of the IEEE “for contributions to algorithms for recognizing objects in images”, and a Fellow of the IAPR “for contributions to computer vision, pattern recognition and biometrics”.  He received a Technical Achievement Award from the IEEE Computer Society “for pioneering contributions to the science and engineering of biometrics”, and both the Meritorious Service Award and the Leadership Award from the IEEE Biometrics Council.  Professor Bowyer served as Editor-In-Chief of both the IEEE Transactions on Biometrics, Behavior, and Identity Science and the IEEE Transactions on Pattern Analysis and Machine Intelligence and has served as General Chair or Program Chair of conferences such as Computer Vision and Pattern Recognition, Winter Conference on Applications of Computer Vision, and Face and Gesture Recognition, and is one of the founding General Chairs of the International Joint Conference on Biometrics.
\end{IEEEbiography}

\begin{IEEEbiography}[{\includegraphics[width=1in, height=1.25in, clip, keepaspectratio]{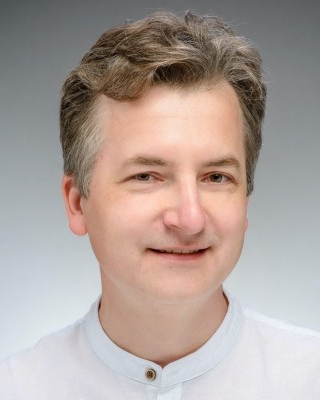}}]{Adam Czajka}
(Senior Member, IEEE) is an Associate Professor at the University of Notre Dame, Notre Dame, IN, USA, where he co-directs the Computer Vision Research Lab and directs the AI Trust and Reliability (AITAR) Lab. His research focuses on human biometrics, especially iris recognition and methods of detecting biometric presentation attacks. In general, Dr. Czajka is interested in a wide spectrum of research in computer vision, pattern recognition, and machine learning, and the non-obvious intersections with psychology, medical sciences, and art. Dr. Czajka is a recipient of the NSF CAREER Award. He serves as the Associate Editor of the {\it IEEE Transactions on Pattern Analysis and Machine Intelligence (T-PAMI)} and {\it IEEE Transactions on Biometrics, Behavior, and Identity Science (T-BIOM)}, is past Senior Associate Editor of the {\it IEEE Transactions on Information Forensics and Security (T-IFS)}, and past Associate Editor of the {\it IEEE Access} and the {\it IEEE Biometrics Compendium}.
\end{IEEEbiography}